%% file: main.tex
\documentclass[10pt,twocolumn,letterpaper]{article}

\usepackage{main/cvpr}              

\input{main/mypackages}

\input{main/mycommands}
\definecolor{cvprblue}{rgb}{0.21,0.49,0.74}
\usepackage[pagebackref,breaklinks,colorlinks,allcolors=cvprblue]{hyperref}

\def\paperID{804}
\def\confName{CVPR}
\def\confYear{2025}

\title{PerLA~$\vcenter{\hbox{\includegraphics[width=0.3in]{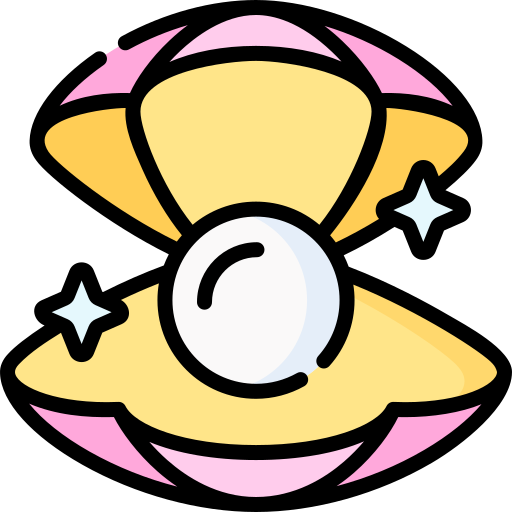}}}$: Perceptive 3D language assistant}

\author{Guofeng Mei$^{1}$ \quad Wei Lin$^2$ \quad Luigi Riz$^1$ \quad Yujiao Wu$^3$ \quad Fabio Poiesi$^1$ \quad Yiming Wang$^1$\\
$^1$Fondazione Bruno Kessler, Italy \quad
$^2$JKU Linz, Austria \quad
$^3$CSIRO, Australia \\
{\tt\small gmei@fbk.eu}
}

\begin{document}

\vspace{-4mm}
\twocolumn[{%
\renewcommand\twocolumn[1][]{#1}%
\maketitle
\begin{center}
    \centering
    \captionsetup{type=figure}
    \includegraphics[width=\linewidth]{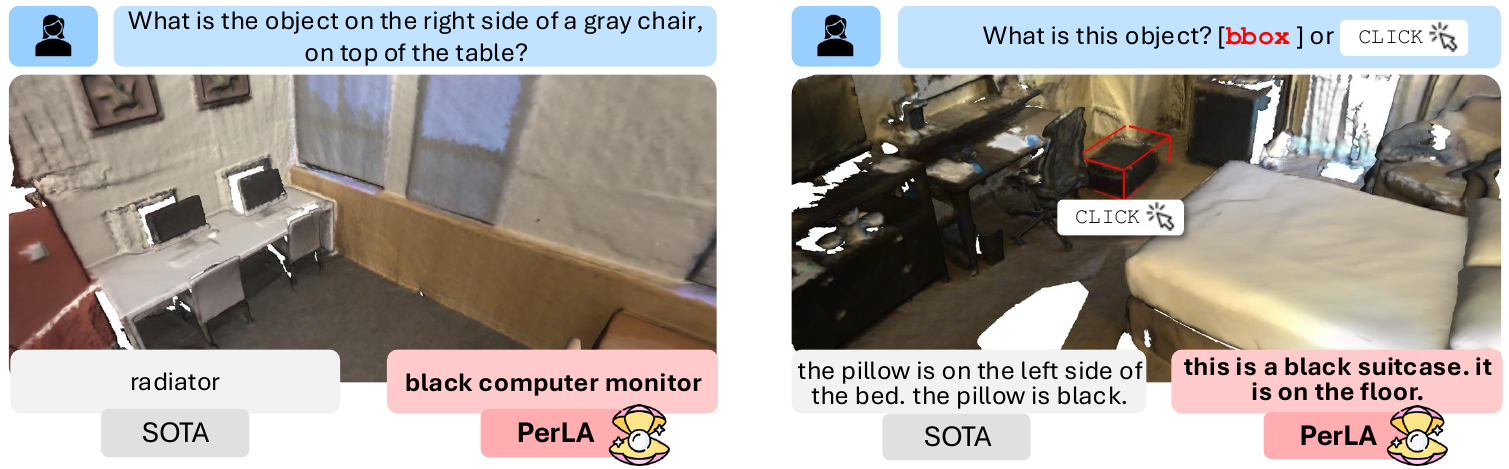}
    \vspace{-8mm}
    \captionof{figure}{\ourmethod is a 3D language assistant that integrates local details with global context to learn informative representations of 3D scenes, whereas state-of-the-art (SOTA) \shortnameplural focus solely on global context information. 
    \ourmethod can provide more accurate responses, correctly distinguishing between objects such as a ``black computer monitor” and a ``black suitcase,” where SOTA models instead fail with hallucinated responses.
    Examples in figures show cases where capturing details from the point cloud matters for accurate output captions.}
\label{fig:teaser}
\end{center}
}]

\input{main/sections/abstract}
\input{main/sections/introduction}

\input{main/sections/related_work}
\input{main/sections/proposed_approach}

\input{main/sections/experiments}

\input{main/sections/conclusions}

\newpage
\bibliographystyle{main/ieeenat_fullname}
\bibliography{main}
\input{main/sections/X_suppl}

\end{document}

%% file: main/mypackages.tex
\usepackage[dvipsnames]{xcolor}
\usepackage{amsmath,amssymb,amsfonts}
\usepackage{multirow,multicol}
\usepackage{natbib}
\usepackage{algorithmic}
\usepackage{graphicx}
\usepackage{textcomp}
\usepackage{xspace}
\usepackage{url,bm}
\usepackage{dsfont}
\usepackage{booktabs}
\usepackage{balance}
\usepackage{soul}
\usepackage{todonotes}
\usepackage{colortbl}
\usepackage{overpic}
\usepackage{enumitem}
\usepackage{nccmath}
\usepackage{courier}
\usepackage{array}
\usepackage{tabularray}
\usepackage{pifont}
\usepackage[accsupp]{axessibility} 
\usepackage{algorithm}
\usepackage{algorithmic}

%% file: main/mycommands.tex
\newcommand{\myparagraph}[1]{\vspace{2pt}\noindent{\bf #1}}

\definecolor{light-gray}{gray}{0.5}
\definecolor{pretty-blue}{RGB}{0, 113, 188}
\def\ie{\emph{i.e.}}
\definecolor{linecolor}{gray}{.895} 
\definecolor{tabhead}{RGB}{213,232,212}
\definecolor{SupMet}{RGB}{218,232,252}
\newcommand{\relativeimpP}[1]{\small{\textcolor{Green}{+#1}}}

\newcommand{\relativeimpZ}[1]{\small{0.0}}

\definecolor{lightblue}{RGB}{218,232,252}
\definecolor{lightred}{RGB}{248,206,204}
\definecolor{lightgreen}{RGB}{213,232,212}

\newcommand{\shortnameplural}{3DLAs\xspace}
\newcommand{\shortname}{3DLA\xspace}
\newcommand{\ourmethod}{PerLA\xspace}
\newcommand{\knn}{$k$-NN\xspace}

\newcommand{\suppmat}{Supp.~Mat.\xspace}

%% file: main/sections/abstract.tex
\vspace{-5mm}
\begin{abstract}
Enabling Large Language Models (LLMs) to understand the 3D physical world is an emerging yet challenging research direction. 
Current strategies for processing point clouds typically downsample the scene or divide it into smaller parts for separate analysis. 
However, both approaches risk losing key local details or global contextual information.
In this paper, we introduce \ourmethod, a 3D language assistant designed to be more perceptive to both details and context, making visual representations more informative for the LLM.
\ourmethod captures high-resolution (local) details in parallel from different point cloud areas and integrates them with (global) context obtained from a lower-resolution whole point cloud.
We present a novel algorithm that preserves point cloud locality through the Hilbert curve and effectively aggregates local-to-global information via cross-attention and a graph neural network.
Lastly, we introduce a novel loss for local representation consensus to promote training stability.
\ourmethod outperforms state-of-the-art 3D language assistants, with gains of up to +1.34 CiDEr on ScanQA for question answering, and +4.22 on ScanRefer and +3.88 on Nr3D for dense captioning. 
Project page: \url{https://gfmei.github.io/PerLA}
\end{abstract}

%% file: main/sections/introduction.tex
\vspace{-7mm}
\section{Introduction}\label{sec:introduction}
3D language assistants (\shortnameplural) jointly process natural language and 3D data to achieve 3D scene understanding, such as recognizing object categories, locations, appearances, and relationships, without requiring specialized models for each recognition task~\cite{hong20233dllm, chen2024ll3da,chen2024grounded3dllm}.
These capabilities are primarily powered by Large Language Models (LLMs) trained in large text corpora~\cite{touvron2023llama}.
These approaches can aggregate multi-view features \cite{hong20233dllm} or process point clouds \cite{chen2024ll3da} to generate 3D representations, which are in turn converted to tokens for the LLM~\cite{zhang2022opt,touvron2023llama}.
However, extracting multi-view representations is computationally costly and often fails to capture essential geometric properties~\cite{hong20233dllm}. 
Directly processing point clouds can yield more accurate results, yet it is even more computationally costly than handling multi-view data, as point clouds typically have rather large cardinalities~\cite{wu2024point}. 
To address this, the point cloud cardinality can be reduced via downsampling~\cite{chen2024ll3da,chen2023end}.
However, as with images~\cite{liu2024llava}, downsampling can compromise downstream task performance due to the reduced model’s ability to perceive fine details of 3D scenes \cite{qiu2021semantic}. 

\cref{fig:teaser} illustrates two cases where our approach can accurately capture details and describe small objects within large scenes, while a state-of-the-art method hallucinates object details~\cite{chen2024ll3da}.
Although fine-grained attributes are critical for performance, directly extracting detailed information from high-resolution 3D data for \shortnameplural remains underexplored.

In this work, we aim to enhance \shortnameplural' ability to perceive finer details in point clouds in order to execute downstream tasks more accurately.
While increasing the number of visual tokens as the point cloud grows in size seems the straightforward solution, our empirical study (\cref{tab:ab_branch}) shows that this solution has limited effectiveness in capturing scene details and it just increases computational burden.
To address this, we propose \ourmethod, a novel \shortname with a perceptive 3D scene encoder that captures detailed point cloud information, allowing the language model to generate \textit{more accurate responses without processing additional tokens}.
\ourmethod first divides the complete 3D scene into non-overlapping local parts to be processed in parallel, then integrates this local information with the global context obtained from a lower-resolution representation of the entire point cloud.
Although dividing visual input has gained popularity in image processing \cite{liu2024llavanext}, it has not yet been applied to point clouds. 
Processing point clouds presents unique challenges beyond those of images, as it requires handling an unordered set of points rather than rasterized pixels.
To address these challenges, we serialize and partition the point cloud before encoding using a Hilbert curve approach~\cite{sagan1994hilbert}, which efficiently preserves locality.
We then combine local and global information through an efficient Hilbert curve-based \knn search and aggregate this information via a novel cross-attention module and Graph Convolutional Network (GCN) to generate highly informative point-level representations for the LLM.
Lastly, we train \ourmethod with a novel loss function designed to promote consensus on local representation, addressing the issue of divergent representations during local-to-global aggregation.
We validate our approach on the question answering benchmark ScanQA~\cite{azuma2022scanqa}, and the 3D dense captioning benchmarks ScanRefer~\cite{chen2020scanrefer} and Nr3D~\cite{achlioptas2020referit3d} to demonstrate the effectiveness of \ourmethod.
\ourmethod demonstrates high transferability on both tasks of 3D question answering and 3D dense captioning, achieving state-of-the-art performance.
In summary, our contributions are:
\begin{itemize}
\item We present a novel perceptive 3D encoder to preserve local and global information for 3D language assistant.
\item We introduce an efficient approach based on Hilbert curve \knn search and cross attention to aggregate local and global information at point level.
\item We propose a novel algorithm based on Graph Neural Network to refine and enhance aggregated information.
\item We introduce a novel loss objective to enable local representation consensus for local-to-global information aggregation.
\end{itemize}

%% file: main/sections/related_work.tex
\section{Related work}\label{sec:related_work}

\myparagraph{3D language understanding} involves understanding 3D scenes by describing or answering to scene-relevant questions in the format of natural language.
Examples of typical downstream tasks include 
\textit{3D Dense Captioning}~\cite{chen2023end,chen2021scan2cap,wang2022spatiality}, 
and \textit{3D Question Answering}~\cite{azuma2022scanqa,ma2022sqa3d,ye20213d,zhao2022toward}.
\textit{3D Dense Captioning} consists of 3D localization of object instances, and textual descriptions of each object instance~\cite{chen2021scan2cap,cai20223djcg,chen2023end, chen2024vote2cap}. 
\textit{3D Question Answering} requires the model equipped with a language decoder to answer questions regarding the visual context in the given 3D scene~\cite{azuma2022scanqa,parelli2023clip}.
Several works focus on addressing the problems of 3D-language pre-alignment~\cite{chen2023unit3d, jin2023context}, or designing adapter layers \cite{hong20233dllm,chen2024ll3da}, or constructing 3D synthetic data \cite{yang20243dgrand}.
In contrast, our work focuses on designing an encoding approach that can enhance 3D-language understanding by capturing fine-grained details.

\myparagraph{3D language assistants (3DLAs)} can be roughly categorized into two types: 
object-level 3DLAs~\cite{qi2025shapellm,tang2024minigpt,xu2023pointllm,huang2023embodied} and 
scene-level 3DLAs~\cite{hong20233dllm,zhu2024llava3d,wang2023chat,chen2024ll3da}.
Object-level 3DLAs learn from large 3D object datasets~\cite{deitke2024objaverse} to connect object-level 3D representation with language models. 
However, they underperform in compositional reasoning in complex 3D scenes with numerous objects. 
Scene-level 3DLAs, such as 3D-LLM \cite{hong20233dllm}, LLaVA-3D~\cite{zhu2024llava3d}, Chat-3D \cite{wang2023chat}, and LL3DA \cite{chen2024ll3da} enable scene understanding via the interaction with objects. 
Because 3D data is orders of magnitude less than 2D data, existing 3DLAs address such lack of data by leveraging pre-trained 2D Large Multimodal Models (LMMs)~\cite{hong20233dllm,zhu2024llava3d}, or data-efficient training recipes~\cite{wang2023chat,chen2024ll3da}. 
3D-LLM \cite{hong20233dllm} leverages 2D pre-trained representations of rendered multi-view images to construct 3D representations and 2D VLMs as backbones.
LLaVA-3D~\cite{zhu2024llava3d} adapts LLaVA~\cite{liu2024llava} for 3D scene understanding by associating 2D patch representations with their positions in 3D space.
Chat-Scene~\cite{huang2024chatscene} improves the referencing and grounding capability and models scene representations as a sequence of object-level representations.
LL3DA \cite{chen2024ll3da} extracts point-level representations from a downsampled 3D scene, and includes both interaction prompts and textual instructions to resemble human interactions with the 3D environment.
Orthogonal to the efforts of aligning 3D-language with limited 3D data in existing \shortnameplural, \ourmethod aims to improve the capability of \shortnameplural in perceiving scene details.

\myparagraph{Visual perception enhancement on multimodal models.} 
Multi-granularity representation learning has been explored in 2D multimodal models~\cite{yang2021focal,fan2021multiscale,lee2022mpvit,chen2021crossvit,ryali2023hiera,shi2025we,gupta2024xt}, showing that combining local and global views yields more informative representations than relying on a single global view.
Mini-Gemini~\cite{li2024mini} introduces a dual vision encoder setup using CLIP ViT~\cite{clip} and ConvNeXt~\cite{liu2022convnet} to process low- and high-resolution views of an input image. 
Models in the LLaVA-Next series~\cite{li2024llava,liu2024llavanext,xu2024llava} and InternLM-XComposer2~\cite{dong2024internlm} use an additional branch to handle view partitions, increasing the number of visual tokens and preserving more fine-grained visual details compared to language-vision models with a single global view branch.
In the 3D domain, Scene-LLM~\cite{fu2024scene} and Segment3D~\cite{huang2025segment3d} improve segmentation accuracy by transferring semantic details from multi-view images to point clouds. 
While extensive work has been done to enhance perception in 2D models and transfer multi-view image information into 3D, methods for preserving detailed representations from point clouds in \shortnameplural remain underexplored. 
Our method aims to address this gap by preserving both global and local visual information to enhance 3D perception.

%% file: main/sections/proposed_approach.tex
\vspace{-2mm}
\section{Perceptive 3D language assistant (\ourmethod)}\label{sec:proposed_approach}
\begin{figure*}[t]
    \centering
    \includegraphics[width=\linewidth]{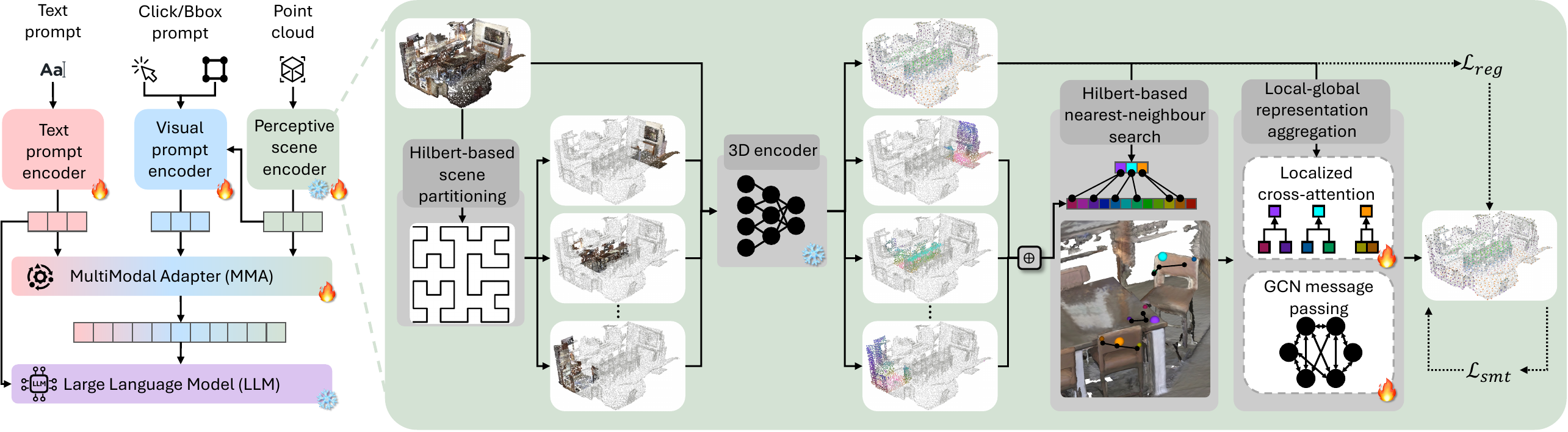}
    \vspace{-4mm}
    \caption{Overview of \ourmethod. (Left): The overall pipeline of \ourmethod, which begins by extracting interaction-aware 3D scene representations. These representations are then projected onto the prefix of textual instructions via MMA, serving as input to a frozen language model (LLM). (Right): The detailed design of \ourmethod. First, the 3D scene is divided into spatially compact regions using Hilbert-based scene serialization~\cite{sagan1994hilbert}. Next, an efficient \knn algorithm associates each point-level global representation with its detail-enriched local representations, creating a comprehensive scene representation through a Graph Convolutional Network (GCN). Finally, smoothness and regularization losses are applied to promote stable learning for the proposed perceptive scene encoder.
    }
    \label{fig:method}
    \vspace{-3mm}
\end{figure*}
\ourmethod takes as inputs \textit{i)} a text prompt in natural language, \textit{ii)} the 3D scene in the form of a point cloud, and \textit{iii)} a visual prompt provided as either a user click or a bounding box. 
The text prompt is processed by a \textit{text prompt encoder} to produce text representations, which are then input to both the \textit{Large Language Model} (LLM) and the \textit{multimodal adapter} (MMA).
The text encoder is a transformer based on BLIP-2~\cite{li2023blip}.
The point cloud is processed by our \textit{perceptive scene encoder}, which generates scene representations that feed into both the MMA and the subsequent encoder.
We will detail the perceptive scene encoder in the next sections.
The visual prompt is processed by the \textit{visual prompt encoder}, which, by combining the perceptive scene encoder’s representations, outputs scene representations that are further processed by the MMA.
For more details on visual prompts, please refer to \suppmat
The MMA takes as input these multimodal representations and outputs tokens for the LLM.
The MMA is implemented as a Q-former~\cite{li2023blip}.
MMA's output is projected into the LLM's representation space through linear projector.
Lastly, these projected representations are processed by the LLM to generate the output response.
We train \ourmethod using data provided by \cite{hong20233dllm} and finetune it for each downstream task.
Fig.~\ref{fig:method} illustrates our approach.

Formally, let $\mathcal{P} {=} \{\left(\bm{p}_i{\in} \mathbb{R}^3, \bm{f}_i {\in} \mathbb{R}^{d_0}\right) \mid i = 1, 2, \dots, N\}$ denote \ourmethod's input point cloud, where 
$\bm{p}_i$ is a point coordinate, 
$\bm{f}_i$ is a $d_0$-dimensional feature vector (e.g., color or normal vector) corresponding to $\bm{p}_i$, and 
$N$ is the number of points (cardinality).
Let $\mathcal{I}^t$ and $\mathcal{I}^v$ denote the input text prompt and visual prompt, respectively.
Based on these inputs, \ourmethod (denoted as $\Phi$) generates a free-form natural language response $\mathcal{O}^t$ for various 3D-related tasks.

\subsection{Perceptive scene encoder}\label{sec:scene_encoder}

To compute point-level representations through the scene encoder, $\mathcal{P}$ is typically downsampled into super-points using Farthest Point Sampling (FPS) \cite{chen2024ll3da,chen2023end}.
However, downsampling and using only $\mathcal{P}$ may hinder the encoder to capture fine-grained details, thus affecting performance in downstream 3D scene understanding tasks.
Our proposed perceptive scene encoder is designed to preserve such scene details without increasing the number of tokens or the representation dimensions of the 3D scene. 
We propose to split $\mathcal{P}$ into parts, and employ a pre-trained 3D scene encoder to encode these parts and the whole point cloud separately. 
We then aggregate the output representations from these different pieces into a single (highly-informative) representation using a cross-attention-based module and Graph Convolutional Network (GCN).

\myparagraph{Hilbert-based scene partitioning.}
We partition $\mathcal{P}$ into $L$ equally-sized parts, each containing the same number of points, $\left \lfloor \frac{N}{L} \right \rfloor$, where $\left \lfloor \cdot \right \rfloor$ denotes the greatest integer less than or equal to its argument. 
Enforcing equal cardinality across parts allows for spatially smaller parts in highly structured regions (areas with more semantic information) and spatially larger parts in less structured regions (areas with less semantic information). 
To achieve this, we adopt a Hilbert curve approach~\cite{sagan1994hilbert} to efficiently serialize the point cloud and partition it into parts~\cite{wu2024point,wang2023octformer,liang2024pointmamba}.


\myparagraph{Global and partial scene encoding.}
Both full-scene point cloud and the partial-scene point clouds are encoded separately by the same pre-trained 3D scene encoder $\phi$.
The 3D scene encoder $\phi$ downsamples the input point cloud into super-points using Farthest Point Sampling (FPS) \cite{moenning2003fast} and produces a representation for each super-point. 
The representation of the full scene, \ie, the global representation, encode the overall scene context, while the representations of the partial scenes, \ie, the local representations, encode scene details.
Let $\mathcal{P}^g = \{\bm{p}^g_i \in \mathbb{R}^3 \}_{i=1}^M$ denote the downsampled full-scene point cloud composed of $M$ points, and $\mathcal{F}^g = \{\bm{f}^g_i \in \mathbb{R}^d \}_{i=1}^M$ the associated global representations.
Let $\mathcal{P}^l = \{\bm{p}^l_i \in \mathbb{R}^3 \}_{i=1}^{L \cdot M}$ denote the set of downsampled partial-scene point clouds, and $\mathcal{F}^l = \{\bm{f}^l_i \in \mathbb{R}^d \}^{L \cdot M}$ the associated local representations.
Note that, we aggregate all the points of the downsampled partial-scene point clouds into the single set $\mathcal{P}^l$, and both the full-scene and the partial-scenes are downsampled to the same number of super-points $M$.
The representations $\mathcal{F}^l$ provide a more detailed view of the scene, as they are derived from the same number of points downsampled from smaller, localized regions, resulting in higher resolution compared to the global versions. 
Yet, their semantic visibility is focused within each local part.
In the following steps, we aggregate local and global information, to enrich the representations with details within the scene context.

\myparagraph{Hilbert-based nearest-neighbor search.}
In order to enhance global representations through local representations, we need to first find the correspondences among them, \ie, which super-points of partial-scene point clouds are spatially neighbors with super-points of the full-scene point cloud.
This can be done via \knn search.
However, traditional \knn searches can be computationally intensive on large-scale or high-resolution point clouds and cannot ensure that the identified neighbors maintain geometric consistency.
To address these limitations, we approximate \knn with an efficient neighbor mapping technique using Hilbert serialized point clouds, which improves query speed while reducing computational complexity. 
With serialized point clouds, we can leverage point indices to perform \knn searches with $O(1)$ complexity~\cite{wu2024point}.

Specifically, we first serialize the union $\mathcal{P}^g \bigcup\mathcal{P}^l$ using the Hilbert curve ordering. We then apply geometric partitioning~\cite{landrieu2018large} on the original point clouds to generate geometric labels $\mathcal{Y}^g \bigcup\mathcal{Y}^l$. The center points of these labels are incorporated as high-order bits (by bit shifting) in the serialized index, ensuring that points with different labels occupy non-overlapping index ranges.
Next, for each $\bm{p}^g_i \in \mathcal{P}^g$, we identify the \texttt{k} nearest super-points within $\mathcal{P}^l$ based on the serialized indexes.
These geometric labels guarantee that $\bm{p}^g_i$ and its nearest local super‐points all originate from the same instance.
Let $\mathcal{P}_{\texttt{k}_i}^l$ denote the set of \texttt{k} nearest local super-points to the global super-point $\bm{p}^g_i$.
Consequently, $\mathcal{F}_{\texttt{k}_i}^l$ are the representations of the super-points in $\mathcal{P}_{\texttt{k}_i}^l$.

\myparagraph{Local-global representation aggregation.}
We make global representations more informative by combining them with information from local representations.
To enable this, we present a novel two-step aggregation technique.
In the first step, we employ a novel cross-attention algorithm between local and global representations using neighborhood information to update global representations, which we term as \textit{localized cross-attention}.
In the second step, we refine global representations through a Graph Convolutional Network based formulation based on message passing, which we term as \textit{GCN message passing} (Fig.~\ref{fig:method}).

\textit{Localized cross-attention} operates with a learning-based weighting mechanism that is a function of the representations and relative positions of global super-points and their associated local neighbors.   
We constrain cross-attention to local neighborhood regions because
\textit{i)} limiting the number of points reduces computational complexity, and 
\textit{ii)} neighbors of a given point are likely to belong to the same object.
Specifically, for each point $\bm{p}^g_i$ and its nearest neighbors $\mathcal{P}_{\texttt{k}_i}^l$, their relative position embeddings $\mathcal{R}_i$ are extracted as
\vspace{-2mm}
\begin{equation}
\vspace{-1mm}
\mathcal{R}_i = \left\{\mathcal{R}_{ij}=\text{pos}\left(\left(\bm{p}^g_i - \bm{p}^l_j\right)/{\sigma}\right) \in \mathbb{R}^d \right\}_{j=1}^{K^l}, 
\end{equation}
where $\sigma > 0$ is a learnable parameter that controls the relative position scaling, and $\text{pos}(\cdot)$ is the 3D Fourier positional embedding~\cite{tancik2020fourier} operation, which satisfies:
\vspace{-2mm}
\begin{equation}
\vspace{-2mm}
\text{pos}(\bm{x}) = 
\left[
\sin\left(2\pi \bm{x} \cdot B \right); 
\cos\left(2\pi \bm{x} \cdot B \right)
\right],
\end{equation}
where $B \in \mathbb{R}^{3 \times (d/2)}$ is a learnable matrix.
We update the global representations through cross-attention as follows
\vspace{-2mm}
\begin{equation}
    \begin{aligned}
       & \bm{s}_{ij} = \left(W_q{\bm{f}^g_i}\right)^\top\left(W_k\left(\bm{f}^l_j+W_r\mathcal{R}_{ij}\right)\right) \big/ \sqrt{d}, \\
       & \bm{s}_i = \{\bm{s}_{ij}\}_{i=1}^{K^l},\bm{w}_{i} = \text{softmax}\left(\bm{s}_i\right), \\
       & \hat{\bm{f}}^g_i = \bm{f}^g_i + \bm{w}_{i}\left(W_v\left(\mathcal{F}^l_{\mathcal{K}_i} + \mathcal{R}_{i}\right)\right),
    \end{aligned}
\end{equation}
where $\hat{\bm{f}}^g_i$ are the updated representations, and $W_q, W_k, W_v, W_r$ are projection parameter matrices.
Let $\hat{\mathcal{F}}^g=\{\hat{\bm{f}}^g_i\}_{i=1}^M$ denote the set of $\hat{\bm{f}}^g_i$.

\textit{GCN message passing} further refines $\hat{\mathcal{F}}^g$ by aggregating information from neighboring global super-points.
Let $\mathcal{G}^g = \{\mathcal{P}^g, \mathcal{W}^g\}$ be a \knn graph, where $\mathcal{P}^g$ is the set of the vertices, \ie, global super-points, and $\mathcal{W}^g \in \mathbb{R}^{N \times N}$ is the adjacency matrix defining the edges.
The construction of the adjacency matrix uses the similarity between super-point representations $\mathcal{F}^G$.
Note that we use the (original) global representations to define the adjacency matrix because we empirically experienced more stability in training as opposed to using the updated global representation, likely due to the fact that they change value during training.
The adjacency matrix $\mathcal{W}^g$ with elements $\mathcal{W}^g_{ij}$ is defined as
\vspace{-2mm}
\begin{equation*}
        \mathcal{W}^g_{ij} {=} 
    \begin{cases}
        \frac{{\bm{f}^g_i}^\top \bm{f}^g_j}{\|\bm{f}^g_i\| \|\bm{f}^g_j\|} \cdot \left({\bm{f}^g_i}^\top \bm{f}^g_j{>}0\right), & \text{if } p^g_j {\in} \mbox{\knn}(p^g_i), \\
        0, & \text{otherwise}.
    \end{cases}
    \vspace{-2mm}
\end{equation*}
Lastly, we perform one-time GCN message passing (MP) on the graph $\mathcal{G}^g$ to obtain an updated feature matrix $\hat{\mathcal{F}}^g$.

Let $\tilde{\mathcal{W}}^g$ be the symmetrically normalized adjacency matrix of $ \mathcal{W}^g$ by $\tilde{\mathcal{W}}^g = \mathbf{D}^{-\frac{1}{2}} \mathcal{W}^g \mathbf{D}^{-\frac{1}{2}} \in \mathbb{R}^{M \times M}$
where \(\mathbf{D}\) is the diagonal degree matrix of $\mathcal{W}^g$. 
\vspace{-2mm}
\begin{equation}
    \hat{\mathcal{F}}^g \leftarrow \text{ReLU}\left(W_m\left(\tilde{\mathcal{W}}^g \hat{\mathcal{F}}^g\right) + W_s\hat{\mathcal{F}}^g\right),
    \vspace{-2mm}
\end{equation}
where $W_m$ and $W_s$ are trainable parameters.
This MP layer structure allows each node to aggregate information from its neighbors while also preserving its own feature representation (via the skip connection).

\subsection{Learning with local representation consensus}\label{sec:learning}

The main learning objective encourages the model to generate outputs that are close to the target responses in the training dataset. 
This objective is typically expressed as a per-token cross-entropy loss $\mathcal{L}_{pred}$~\cite{chen2024ll3da,hong20233dllm} (see details in our \suppmat).
We found that adding a consensus term to $\mathcal{L}_{pred}$ that encourages local regularization of representations improves training stability.
Firstly, points belonging to the same object (\ie, local neighborhoods) should share similar aggregated representations. 
Secondly, aggregated super-point representations should remain close to their original point representations to preserve context knowledge. 
Our novel consensus loss $\mathcal{L}_{\text{con}}$ is defined as
\vspace{-2mm}
\begin{equation}\label{eq:agg_loss}
\vspace{-2mm}
\begin{aligned}
        \mathcal{L}_{\text{con}} & = \mathcal{L}_{\text{smt}} + \mu\mathcal{L}_{\text{reg}}, \\
        \mathcal{L}_{\text{smt}} &= \sum_{i=1}^{N}\sum_{j=1}^{N}\mathcal{W}^g_{ij} \left\|\frac{\hat{\bm{f}}^g_i}{\sqrt{d_{i}}} {-} \frac{\hat{\bm{f}}^g_j}{\sqrt{d_j}}\right\|, \\
        \mathcal{L}_{\text{reg}} &= \sum^{N}_{i=1} \left\|\hat{\bm{f}}^g_i - \bm{f}^g_i\right\|,
\end{aligned}
\end{equation}
%
where $d_k = \sum_l \mathcal{W}^g_{kl}$ is the diagonal matrix representing row-wise sums of $\mathcal{W}^g$.
The first term $\mathcal{L}_{\text{smt}}$ enforces spatial connectivity by promoting similarity among fused representations of points within the same object, addressing the first attribute. 
$\mathcal{L}_{\text{reg}}$ is a regularization term that promotes the learned representations stay close to the original global representations.
$\mu$ is a hyperparameter.
The overall loss $\mathcal{L}$, is a weighted sum of $\mathcal{L}_{pred}$ and $\mathcal{L}_{con}$
\vspace{-2mm}
\begin{equation}
    \mathcal{L} = \lambda {\mathcal{L}}_{con} + {\mathcal{L}}_{pred},
    \vspace{-2mm}
\end{equation}
where $\lambda$ is the balance hyperparameter.
By jointly training with these two loss terms, we encourage the model to learn more object-aware representations and enhance its ability to extract finer details. 
This approach also preserves global context information, which is essential for capturing spatial relationships between different objects.

During inference, we use beam search to predict response $\mathcal{O}^t$ that maximizes the following objective:
\begin{equation}
    \mathcal{O}^t = \arg\max_{\mathcal{O}}\Phi\left(\mathcal{O}\mid\mathcal{P},\mathcal{I}^t,\mathcal{I}^v\right).
    \vspace{-2mm}
\end{equation}
where we set a beam size of 4.

%% file: main/sections/experiments.tex
\section{Experiments}\label{sec:experiment}
We evaluate \ourmethod on 3D Question Answering and 3D Dense Captioning downstream tasks, and compare its performance with state-of-the-art methods from the literature.

\myparagraph{Datasets.} We conduct experiments using the ScanNet dataset~\cite{dai2017scannet}, which encompasses 1,201 training and 312 validation scenes, featuring diverse and complex indoor 3D environments. Language annotations are from ScanQA~\cite{azuma2022scanqa},  ScanRefer~\cite{chen2020scanrefer}, Nr3D~\cite{achlioptas2020referit3d}, and the ScanNet subset of 3D-LLM~\cite{hong20233dllm}, collectively supporting a range of tasks including instance and scene descriptions, conversations, embodied planning, and question answering. 
For additional data statistics, please refer to the supplementary materials.

\myparagraph{Metrics.} 
We follow LL3DA's evaluation protocol~\cite{chen2024ll3da} to evaluate the quality of output responses. 
We use the abbreviations C, B4, M and R for CiDEr~\cite{vedantam2015cider}, BLEU-4~\cite{papineni2002bleu}, METEOR~\cite{banerjee2005meteor}, and Rouge-L~\cite{lin2004rouge}, respectively.

\myparagraph{Implementation details.} As in~\cite{chen2024ll3da,chen2021scan2cap}, we input 40,000 randomly sampled points from each 3D scene. 
We divide the point cloud into partitions $L{=}6$ and choose $\texttt{k}{=}4L{=}24$ neighbors to enhance global representations. 
We set $\lambda{=}\mu{=}0.1$ in the loss.
We use the pre-trained OPT-1.3B~\cite{zhang2022opt} language model, kept frozen and loaded in \texttt{float16} precision.
We use the AdamW~\cite{loshchilov2017fixing} optimizer with a weight decay of 0.1, applying a cosine annealing scheduler that decays the learning rate from \(10^{-4}\) to \(10^{-6}\) over approximately 100,000 iterations. 
All tasks are trained with a total batch size of 16. 
Each training process is completed within two days using up to two NVIDIA H100 P0 (96GB) GPUs.
For each evaluation, we fine-tune the model's parameters on the respective task for about 30,000 iterations.

\subsection{Results}\label{sec:results}

\input{main/tables/tab_qa}
\begin{figure*}[t]
    \centering
    \includegraphics[width=1\linewidth]{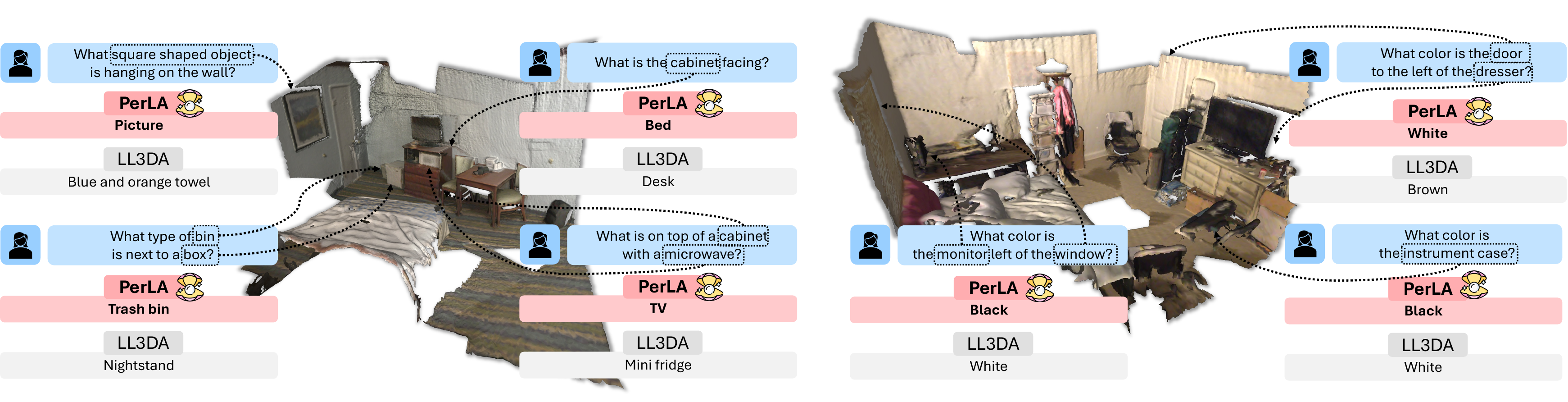}
    \vspace{-8mm}
    \caption{The qualitative comparison between our method, \ourmethod, and LL3DA~\cite{chen2024ll3da} on the ScanQA~\cite{azuma2022scanqa} dataset shows that our approach achieves higher accuracy in responding to ``what''-related questions.
    }
    \label{fig:qa}
    \vspace{-5mm}
\end{figure*}

\myparagraph{3D question answering} is a task that involves answering questions about a 3D scene. It allows a model to provide information about the objects, relationships, and attributes within a 3D environment based on a given question. 
\cref{tab:scanqa} shows the results of ScanQA's validation and test sets.
Classification-based methods (CLS) select responses from a predefined answer set. 
Generation-based approaches (GEN) generate the entire textual response. 
\ourmethod consistently outperforms existing approaches across all evaluation sets and metrics, in particular with +1.34 CiDEr score over LL3DA.
We also compare against our reproduced version of LL3DA, where \ourmethod scores +3.76 CiDEr.

\cref{fig:qa} provides qualitative comparisons between LL3DA and \ourmethod on the ScanQA benchmark~\cite{azuma2022scanqa}. 
It highlights \ourmethod's accuracy in answering questions regarding object attributes and spatial relationships within a 3D scene. 
\ourmethod provides precise answers, correctly identifying objects such as a``picture", ``trash bin", and ``TV" along with their specific colors and types, while LL3DA often yields incorrect or less specific responses.
In particular, \ourmethod accurately identifies the colors of objects within the scene, such as the door, the monitor, and the instrument case, correctly answering the questions about these attributes. In contrast, LL3DA frequently misidentifies colors, labeling objects as ``White" instead of ``Black," for example. This illustrates \ourmethod's superior ability to capture fine-grained details and contextual information, delivering more accurate answers in complex environments.
\input{main/tables/tab_refer_nr3d}
\begin{figure*}[t]
    \centering
    \includegraphics[width=1\linewidth]{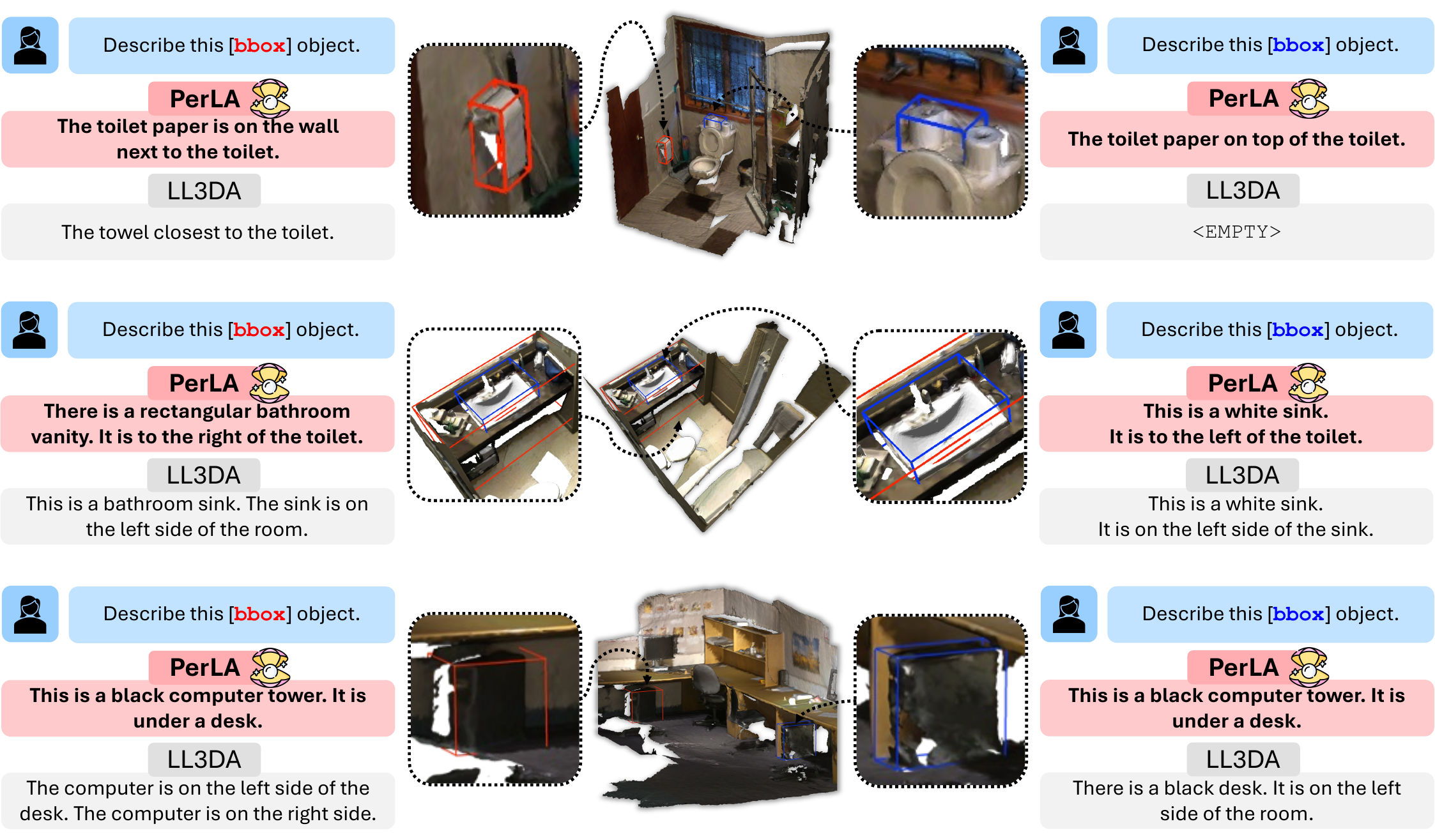}
    \put(-300,195){\color{black}\footnotesize{(a) 3D visual dense captioning on Nr3D~\cite{achlioptas2020referit3d}}}
    \put(-300,100){\color{black}\footnotesize{(b) 3D visual dense captioning on ScanRefer~\cite{chen2020scanrefer}}}
    \put(-300,5){\color{black}\footnotesize{(c) 3D visual dense captioning on ScanRefer~\cite{chen2020scanrefer}}}
    \vspace{-4mm}
    \caption{Qualitative comparisons on the dense captioning task across the Nr3D~\cite{achlioptas2020referit3d} and ScanRefer~\cite{chen2020scanrefer}. We compare the results of our \ourmethod with LL3DA~\cite{chen2024ll3da}. \ourmethod generates accurate descriptions, effectively capturing fine-grained object attributes and spatial relationships.}
    \label{fig:caption}
    \vspace{-3mm}
\end{figure*}

\myparagraph{3D dense captioning} involves localizing and describing each 3D instance within complex 3D environments.
\cref{tab:3D_dense_captioning} shows the results on ScanRefer~\cite{chen2020scanrefer} and Nr3D~\cite{achlioptas2020referit3d} benchmarks. 
Following previous works~\cite{hong20233dllm,chen2024ll3da},
we use the m@$k$IoU metric, where $m \in \{\text{C, B4, M, R}\}$ and \( k \) denote the IoU threshold. 
As in \cite{jiao2022more, chen2021scan2cap}, we report C@0.25 and C@0.5 for ScanRefer, and C@0.5 for Nr3D.
UniT3D~\cite{chen2023unit3d}, 3DJCG~\cite{cai20223djcg}, and 3D-VLP~\cite{jin2023context} are pre-trained on multiple 3D vision and language tasks from ScanNet scenes.
UniT3D uses image caption models and multi-view images to generate extra instance captions for pre-training.
For fair comparison, we report results from models trained using standard per-word cross-entropy loss without additional 3D scenes. 
Box annotations are estimated with Vote2CapDETR and used as visual prompts.
\ourmethod consistently outperforms comparison methods on both benchmarks. 
\ourmethod significantly outperforms LL3DA by scoring +3.75 and +4.22 CiDEr on ScanRefer, and +3.88 CiDEr on Nr3D.
These results highlight the effectiveness of \ourmethod in 3D dense captioning tasks.

\cref{fig:caption} provides examples of qualitative results of LL3DA and \ourmethod on Nr3D and ScanRefer.
PerLA produces more accurate and detailed descriptions, effectively capturing spatial relationships and fine-grained attributes, such as the positioning of objects relative to surrounding elements.

\subsection{Ablation study}\label{sec:ab_study}
\input{main/tables/ab_branch}
\myparagraph{Local, global, and GCN representations.}
To investigate the impact of local and global representations, we design two variants of \ourmethod: one using only local representations and the other using only global representations. 
We also report an extended version of LL3DA, termed LL3DA$^\dag$, in which we increase the number of query tokens used to encode local partitions. 
Representations are first extracted from each partition, and FPS is applied to sample 1,024 points with their corresponding representations from the union of these partitions. These representations are then passed through a multimodal adapter to produce 32 global tokens.
LL3DA generates an additional 32 global tokens from the point cloud of the entire scene, resulting in a total of 64 tokens, which are concatenated and fed into the LLM for response generation.
(For details of LL3DA$^\dag$ please refer to the Supp. Mat.).
\cref{tab:ab_branch} shows that the combination of local and global representations yields a much better performance than the two independently used.
\cref{tab:ab_branch} also shows that LL3DA$^\dag$ does not reach \ourmethod performance, underlying the importance of information exchange between local and global representations.
To evaluate the effectiveness of GCN, we also report the results without using GCN (w/o GCN). Compared to our \ourmethod, which incorporates GCN, it consistently improves performance across the three datasets, confirming its effectiveness.

\myparagraph{Localized cross-attention.}
To evaluate the effectiveness of our novel localized cross-attention module, we tested max pooling and mean pooling as alternative methods.
\cref{tab:ab_attn} shows that localized cross-attention significantly outperforms both mean pooling and max pooling across all metrics. 
These results indicate that localized cross-attention enhances ability of model to generate more informative scene representations, thereby improving performance on 3D question answering tasks.
This improvement comes from the incorporation of positional information and semantic similarity in cross-attention, which helps to exclude neighboring points that do not belong to the same object.
\input{main/tables/tab_ab_attn}

\myparagraph{Loss function.} 
To evaluate the effectiveness of our novel loss term \(\mathcal{L}_{\text{con}}\), we train \ourmethod using different combinations of \(\mathcal{L}_{\text{con}}\)'s components (Eq.~\ref{eq:agg_loss}).
Since our task is intended for downstream applications, each combination also includes the task-specific loss $\mathcal{L}_{\text{pre}}$.
\cref{tab:ab_loss} shows that adding \(\mathcal{L}_{\text{smt}}\) to \(\mathcal{L}_{\text{pre}}\) improves all metrics (C, B4, M, and R).
Incorporating the loss of regularization \(\mathcal{L}_{\text{reg}}\) further enhances performance, particularly in metrics C, M, and R. 
These results show the effectiveness of learning jointly with both semantic awareness and regularization losses.
\input{main/tables/tb_ab_loss}

\myparagraph{Number of partitions.}
To evaluate the effectiveness of different number of partitions, we test \ourmethod by splitting the point cloud into 4, 6, and 8 partitions.
\cref{tab:ab_part} shows that increasing the number of partitions generally improves performance, with 6 partitions yielding the best results.
Although 8 partitions outperform 4, they show slightly lower performance than 6. 
This suggests that 6 partitions provide a better balance between granularity and effectiveness, capturing relevant scene details with lower computational complexity.
More ablation studies, including an analysis of performance in different training strategies, refer to \suppmat.

\input{main/tables/tb_ab_part}

%% file: main/tables/tab_qa.tex
\begin{table*}[t]
    \centering
    \small
    \tabcolsep 7pt
    \caption{Comparative results for 3D Question Answering on ScanQA~\cite{azuma2022scanqa} benchmark. 
    CLS and GEN denote classification-based and generation-based methods, respectively. 
    LL3DA (repr.) is results of LL3DA we reproduced. 
    \ourmethod outperforms all the other methods.}
    \label{tab:scanqa}
    \vspace{-2mm}
    \resizebox{\textwidth}{!}{%
    \begin{tabular}{lc|cccc|cccc|cccc}
        \toprule
        \multirow{2}{*}{Method} & \multirow{2}{*}{Type} & \multicolumn{4}{c|}{Validation} & \multicolumn{4}{c|}{Test w/ object} & \multicolumn{4}{c}{Test w/o object} \\
        & & C$\uparrow$ & B4$\uparrow$ & M$\uparrow$ & R$\uparrow$ & C$\uparrow$ & B4$\uparrow$ & M$\uparrow$ & R$\uparrow$ & C$\uparrow$ & B4$\uparrow$ & M$\uparrow$ & R$\uparrow$ \\
        \midrule
        ScanQA\cite{azuma2022scanqa} & CLS & 64.86 & 10.08 & 13.14 & 33.33 & 67.29 & 12.04 & 13.55 & 34.34 & 60.24 & 10.75 & 12.59 & 31.09 \\
        Clip-Guided\cite{parelli2023clip} & - & - & - & - & - & 69.53 & 14.64 & 13.94 & 35.15 & 62.83 & 11.73 & 13.28 & 32.41 \\
        Multi-CLIP\cite{delitzas2023multi} & CLS & - & - & - & - & 68.70 & 12.65 & 13.97 & 35.46 & 63.20 & 12.87 & 13.36 & 32.61 \\
        3D-VLP\cite{jin2023context} & CLS & 66.97 & 11.15 & 13.53 & 34.51 & 70.18 & 11.23 & 14.16 & 35.97 & 63.40 & 15.84 & 13.13 & 31.79 \\
        3D-VisTA\cite{zhu20233d} & - & - & - & - & - & 68.60 & 10.50 & 13.80 & 35.50 & 55.70 & 8.70 & 11.69 & 29.60 \\
        \midrule
        3D-LLM~\cite{hong20233dllm} & GEN & 69.40 & 12.00 & 14.50 & 35.70 & 69.60 & 11.60 & 14.90 & 35.30 & - & - & - & - \\
        LL3DA\cite{chen2024ll3da} & GEN & {76.79} & {13.53} & {15.88} & {37.31} & {78.16} & {13.97} & {16.38} & {38.15} & {70.29} & {12.19} & {14.85} & {35.17} \\
        LL3DA (repr.) & GEN & 74.37 & 13.50 & 15.09 & 36.31 & - & - & - & - & - & - & - & - \\
        \rowcolor{lightgreen}\ourmethod & GEN & \bf78.13 & \bf14.49 & \bf17.44 & \bf39.60  & \bf80.91 & \bf17.21 & \bf16.49 & \bf40.71 & \bf74.82 & \bf14.97 & \bf15.23 & \bf38.18\\
        \midrule
        \rowcolor{linecolor}$\Delta$ w.r.t. LL3DA\cite{chen2024ll3da} & - & \relativeimpP{1.34} & \relativeimpP{0.96} & \relativeimpP{1.56} & \relativeimpP{2.29} & \relativeimpP{2.75} & \relativeimpP{3.24} & \relativeimpP{0.11}& \relativeimpP{2.56} & \relativeimpP{4.53} & \relativeimpP{2.78} & \relativeimpP{0.38} & \relativeimpP{3.01} \\
        \bottomrule
    \end{tabular}
    }
    \vspace{-3mm}
\end{table*}

%% file: main/tables/tab_refer_nr3d.tex
\begin{table*}[t]
\small
\centering
\tabcolsep 7pt
\caption{Comparative results for 3D Dense Captioning on ScanRefer~\cite{chen2020scanrefer} and Nr3D~\cite{achlioptas2020referit3d} benchmarks.
LL3DA (repr.) is results of LL3DA we reproduced.
Generally, \ourmethod outperforms all other methods on both benchmarks.}
\label{tab:3D_dense_captioning}
\vspace{-3mm}
\resizebox{\textwidth}{!}{%
\begin{tabular}{l|cccc|cccc|cccc}
\toprule
\multirow{2}{*}{Method} & \multicolumn{4}{c|}{ScanRefer@0.25} & \multicolumn{4}{c|}{ScanRefer@0.5} & \multicolumn{4}{c}{Nr3D@0.5} \\
& C$\uparrow$ & B4$\uparrow$ & M$\uparrow$ & R$\uparrow$ & C$\uparrow$ & B4$\uparrow$ & M$\uparrow$ & R$\uparrow$ & C$\uparrow$ & B4$\uparrow$ & M$\uparrow$ & R$\uparrow$ \\
\midrule
Scan2Cap\cite{chen2021scan2cap} & 56.82 & 34.18 & 26.29 & 55.27 & 39.08 & 23.32 & 21.97 & 44.78 & 27.47 & 17.24 & 21.80 & 49.06 \\
MORE\cite{jiao2022more} & 62.91 & 36.25 & 26.75 & 56.33 & 40.94 & 22.93 & 21.66 & 44.42 & - & - & - & - \\
SpaCap3D\cite{wang2022spatiality} & - & - & - & - & - & 44.02 & 25.26 & 22.33 & 33.71 & 19.92 & 22.61 & 50.50 \\
REMAN\cite{mao2023complete} & 62.01 & 36.37 & 26.76 & 56.25 & 45.00 & 26.31 & 23.13 & 46.96 & 34.81 & 20.37 & 22.71 & 50.90 \\
D3Net\cite{chen2022d} & - & - & - & - & - & 51.67 & - & - & 35.26 & 20.42 & 22.77 & 53.38 \\
Contextual\cite{zhong2022contextual} & - & - & - & - & - & 46.07 & 23.40 & 23.95 & - & - & - & - \\
UniT3D\cite{chen2023unit3d} & - & - & - & - & 46.69 & 27.52 & 21.91 & 45.98 & - & - & - & - \\
3DJCG\cite{cai20223djcg} & 64.70 & 40.17 & 27.63 & 59.23 & 49.48 & 31.63 & 24.36 & 50.80 & 38.06 & 22.82 & 23.77 & 52.99 \\
3D-VLP\cite{jin2023context} & 70.73 & 41.03 & 28.14 & \bf59.72 & 54.94 & 32.31 & 24.83 & 51.51 & - & - & - & - \\
3D-VisTA*\cite{zhu20233d} & - & - & - & - & 61.60 & 34.10 & 26.80 & 55.00 & - & - & - & - \\
Vote2CapDETR\cite{chen2023end} & 71.45 & 39.34 & 28.25 & 59.63 & 61.81 & 34.46 & 26.22 & 54.40 & 43.84 & 26.68 & 25.41 & 54.43 \\
LL3DA\cite{chen2024ll3da} & {74.17} & {41.41} & {27.76} & {59.53} & {65.19} & {36.79} & 25.97 & {55.06} & {51.18} & {28.75} & {25.91} & {56.61} \\
LL3DA (repr.) & 71.86 & 39.57 & 27.29 & 58.37 & 63.79 & 35.67 & 25.94 & 54.56 & 48.38 & 28.36 & 25.72 & 55.66 \\
\rowcolor{lightgreen}\ourmethod & \bf 77.92 & \bf43.41 & \bf28.97 & 59.69 & \bf69.41 & \bf38.02 & \bf29.07 &  \bf56.80 & \bf55.06 & \bf31.24 & \bf28.52 & \bf59.13 \\
\midrule
\rowcolor{linecolor}$\Delta$ w.r.t. LL3DA\cite{chen2024ll3da}  & \relativeimpP{3.75} & \relativeimpP{2.00} & \relativeimpP{1.21} & \relativeimpP{0.16} & \relativeimpP{4.22} & \relativeimpP{1.23} & \relativeimpP{2.27}& \relativeimpP{1.74} & \relativeimpP{3.88} & \relativeimpP{2.49} & \relativeimpP{2.61} & \relativeimpP{2.52} \\
\bottomrule
\end{tabular}
}
\vspace{-5mm}
\end{table*}

%% file: main/tables/ab_branch.tex
\begin{table*}[!hbt]
\small
\centering
\tabcolsep 8pt
\caption{Ablation study of our local-to-global, GCN representation aggregation algorithms on ScanQA, ScanRefer\cite{chen2020scanrefer} and Nr3D\cite{achlioptas2020referit3d} benchmarks. LL3DA$^\dag$ denotes an extended version of LL3DA with an increased number of query tokens.}
\label{tab:ab_branch}
\vspace{-3mm}
\resizebox{\textwidth}{!}{%
\begin{tabular}{l|cccc|cccc|cccc}
\toprule
\multirow{2}{*}{Method} & \multicolumn{4}{c|}{ScanQA} & \multicolumn{4}{c|}{ScanRefer@0.5} & \multicolumn{4}{c}{Nr3D@0.5} \\
 & C$\uparrow$ & B4$\uparrow$ & M$\uparrow$ & R$\uparrow$ & C$\uparrow$ & B4$\uparrow$ & M$\uparrow$ & R$\uparrow$ & C$\uparrow$ & B4$\uparrow$ & M$\uparrow$ & R$\uparrow$ \\
\midrule
LL3DA$^\dag$ & 74.54 & 12.89 & 15.11 & 36.96 & 62.25 & 34.50 & 25.55 & 53.84 & 48.70 & 28.24 & 25.72 & 55.59 \\
Global & 74.49 & 13.50 & 15.16 & 36.55 & 63.79 & 35.67 & 25.94 & 54.56 & 48.38 & 28.36 & 25.72 & 55.66 \\
Local & 73.55 & 12.98 & 14.95 & 36.07 & 62.79 & 34.78 & 25.94 & 54.16 & 49.09 & 28.20 & 25.80 & 55.89 \\
w/o GCN & {77.61} & {13.87} & {15.93} & {39.28} & {69.04} & {37.43} & 28.71 & {55.59} & {54.30} & {29.78} & {27.41} & {58.32}\\ 
\rowcolor{lightgreen}\ourmethod & \bf78.13 & \bf14.49 & \bf17.44 & \bf39.60 & \bf69.41 & \bf38.02 & \bf29.07 &  \bf56.80 & \bf55.06 & \bf31.24 & \bf28.52 & \bf59.13 \\
\bottomrule
\end{tabular}
}
\vspace{-4mm}
\end{table*}

%% file: main/tables/tab_ab_attn.tex
\begin{table}[t]
    \centering
    \caption{Ablation study on the impact of constrained attention on ScanQA validation dataset.}
    \label{tab:ab_attn}
    \tabcolsep 3pt
    \vspace{-4mm}
    \resizebox{\columnwidth}{!}{%
    \begin{tabular}{cccc|cccc|cccc} 
        \toprule
        \multicolumn{4}{c}{cross-attention} & \multicolumn{4}{c}{mean pooling}  & \multicolumn{4}{c}{max pooling}\\
        \midrule
        C$\uparrow$ & B4$\uparrow$ & M$\uparrow$ & R$\uparrow$ & C$\uparrow$ & B4$\uparrow$ & M$\uparrow$ & R$\uparrow$  & C$\uparrow$ & B4$\uparrow$ & M$\uparrow$ & R$\uparrow$\\ 
        \midrule
        \bf78.1 & \bf14.5 & \bf17.4 & \bf39.6  & 74.4 & 12.5 & 15.2 & 36.5 & 74.0 & 13.0 & 14.8 & 35.7\\ 
        \bottomrule
    \end{tabular}
    }
\end{table}

%% file: main/tables/tb_ab_loss.tex
\begin{table}[t]
    \centering
    \caption{Ablation study on the impact of joint learning by different loss combination on ScanQA validation dataset.}
    \label{tab:ab_loss}
    \vspace{-3mm}
    \tabcolsep 9pt
    \resizebox{\columnwidth}{!}{%
    \begin{tabular}{ccc|cccc} 
        \toprule
        \multicolumn{3}{c}{Loss} & \multicolumn{4}{c}{ ScanQA (Validation)} \\
        \midrule
        $\mathcal{L}_{\text{pre}}$ & $\mathcal{L}_{\text{smt}}$ & $\mathcal{L}_{\text{reg}}$ & C$\uparrow$ & B4$\uparrow$ & M$\uparrow$ & R$\uparrow$\\ 
        \midrule
        \checkmark &            &            & 75.31 & 13.83 & 15.90 & 37.48 \\ 
        \checkmark & \checkmark &            & 76.62 & 13.96 & 16.99 & 38.14\\ 
        \checkmark &            & \checkmark & 76.47 & 14.07 & 16.72 & 38.79\\ 
        \rowcolor{lightgreen}\checkmark & \checkmark & \checkmark & \bf78.13 & \bf14.49 & \bf17.44 & \bf39.60  \\ 
        \bottomrule
    \end{tabular}
    }
    \vspace{-3mm}
\end{table}

%% file: main/tables/tb_ab_part.tex
\begin{table}[t]
    \centering
    \caption{Ablation study on the impact of number of parts on ScanQA validation dataset.}
    \label{tab:ab_part}
    \tabcolsep 3pt
    \vspace{-4mm}
    \resizebox{\columnwidth}{!}{%
    \begin{tabular}{cccc|cccc|cccc} 
        \toprule
        \multicolumn{4}{c}{4} & \multicolumn{4}{c}{6} & \multicolumn{4}{c}{8} \\
        \midrule
        C$\uparrow$ & B4$\uparrow$ & M$\uparrow$ & R$\uparrow$ & C$\uparrow$ & B4$\uparrow$ & M$\uparrow$ & R$\uparrow$  & C$\uparrow$ & B4$\uparrow$ & M$\uparrow$ & R$\uparrow$\\ 
        \midrule
        76.9 & 13.6 & 16.1 & 37.3 & \bf78.1 & \bf14.5 & \bf17.4 & \bf39.6 & 78.0 & 14.5 & 17.2 & 39.4\\ 
        \bottomrule
    \end{tabular}
    }
    \vspace{-3mm}
\end{table}

%% file: main/sections/conclusions.tex
\section{Conclusions}\label{sec:conclusions}
We presented \ourmethod, a perceptive 3D language assistant capable of capturing both detailed and contextual information to enhance visual representations for LLMs. 
\ourmethod features a dual-branch architecture: the global branch processes superpoints from the whole point cloud via downsampling, while the local branch focuses on partitioned regions. 
We demonstrated that by integrating representations from both branches, \ourmethod effectively captures scene details, reducing hallucinations. 
Moreover, we employ a Graph Convolutional Network to facilitate information exchange among neighboring local and global superpoints, and introduce a novel loss term for local representation consensus to promote training stability.
Experiments on the ScanQA, ScanRefer, and Nr3D benchmarks highlight the effectiveness of our approach, setting a new state-of-the-art performance in 3D question answering and dense captioning.

\myparagraph{Limitations \& Future Work.}
\ourmethod primarily focuses on enhancing performance using point cloud input, but integrating it with optimized modules, such as token merging, presents a promising direction to extend its capabilities. Although \ourmethod shows strong performance on standard benchmarks, future work could explore its robustness and generalizability in more complex 3D scenarios, broadening its applicability in diverse real-world settings.

\noindent\textbf{Acknowledgement.} 
This work was supported by PNRR FAIR - Future AI Research (PE00000013) and ICSC National Research Centre for HPC, Big Data and Quantum Computing (CN00000013), funded by NextGeneration EU.

%% file: main/sections/X_suppl.tex
\appendix

\maketitlesupplementary

\renewcommand{\thetable}{\Alph{table}}
\renewcommand{\thefigure}{\Alph{figure}}

\section{Introduction}
In this supplementary material, we begin by thoroughly describing the components of \ourmethod that are distinct from the perspective scene encoder~(\cref{supp:components}). 
Next, we offer a more technical explanation of Hilbert-based serialization and partitioning for point clouds, including detailed algorithms and an analysis of computational efficiency~(\cref{supp:hilbert}).
Moreover, we provide the complete list of datasets that are involved in model training and testing~(\cref{supp:dataset}). 
In addition, we include an extended analysis performed during the submission phase, and offer additional implementation details and deeper discussion of the results~(\cref{supp:analysis}). 
Finally, we present additional interesting qualitative results of \ourmethod~in comparison with state-of-the-art competitors.

\section{More Details of \ourmethod}
\label{supp:components}
In this section, we describe in detail additional components in \ourmethod, including the off-the-shelf 3D encoder within our perceptive scene encoder, the multimodal prompts involved in the 3DLA interaction as well as their corresponding prompt encoders, and the multimodal adapter that integrates both multimodal prompt and the 3D scene representation to form query tokens that are interpretable by the LLM. 
Lastly, we explain the next token prediction loss.

\subsection{3D encoder}
We adopt the same 3D encoder architecture as in LL3DA~\cite{chen2024ll3da} to process the 3D point cloud. This 3D encoder first tokenizes the input into 2,048 point tokens, by uniformly sampling across the input point cloud using a set-abstraction layer~\cite{qi2017pointnet++}. The point tokens are then passed through three cascaded transformer encoder blocks, employing masking radius of 0.16, 0.64, and 1.44, respectively. To further refine the token representation, an additional set-abstraction layer is introduced between the first two transformer blocks, downsampling the tokens to 1,024. The final output of this 3D encoder is a feature matrix of shape $\mathbb{R}^{1,024 \times 256}$, where each of the point tokens is encoded as a 256-dimensional feature vector.

\subsection{Multimodal prompts}

Prompts in \ourmethod are multimodal and are designed to simulate user-driven interactions within 3D environments. Specifically, we consider both visual and textual prompts as in prior work~LL3DA~\cite{chen2024ll3da}. The \textit{visual prompts} involve visual cues coming from the 3D content, \eg, user clicks or bounding boxes around objects, while the \textit{textual prompts} involve user instructions expressed in natural language format. Such multimodal prompts allow \ourmethod{} to interpret and respond effectively to intuitive user inputs. In the following, we present how each type of prompts is encoded.

\noindent\textbf{Scene-aware visual prompt encoder.}
The visual prompt encoder aims to process visual prompts into representations that are then easier to be processed by the LLM. In addition to prior work which applies positional encoding followed by an MLP layer to process the visual prompts (as described in Eq.~3 of \cite{chen2024ll3da}), we further enhance this approach by augmenting the positional encoding with the detail-enriched global scene representations $\hat{\mathcal{F}}^g$, obtained by our perceptive scene encoder. 
Therefore, the visual prompt encoder is more aligned to the scene representation, helping to improve the model performance as empirically proved in \cref{tab:sub_ab_prompt}.

Specifically, each user click is first normalized to a range of $[0, 1]$ based on the dimensions of the input 3D scene, where $ p_{\text{click}} \in \mathbb{R}^3 $. We then encode $ p_{\text{click}} $ using 3D Fourier positional embeddings, denoted as $ \text{pos}(p_{\text{click}}) $.
The box annotation is represented by the ROI feature $f_{\text{box}} \in \mathbb{R}^d$ with the center point $p_{box}$ extracted by a pre-trained 3D object detector~\cite{chen2023end}. 
We first merge the two types of visual prompts with the scene representations regarding the neighborhood points, and then we use an MLP to project the merged representations as follows:
\begin{equation*}
\begin{aligned}
    \bm{f}_{\text{cli}} &= \text{MLP}_{\text{cli}}\left( \text{pos}(p_{\text{cli}}), h\left( \left\{ \hat{\bm{f}}^g_i \mid \bm{p}^l_{j} \in \mathcal{N}(p_{\text{cli}}) \right\} \right) \right), \\
    \bm{f}_{\text{box}} &= \text{MLP}_{\text{box}}\left( f_{\text{box}}, h\left( \left\{ \hat{\bm{f}}^g_i \mid \bm{p}^l_{j} \in \mathcal{N}(p_{\text{box}}) \right\} \right) \right),
\end{aligned}
\end{equation*}
where $h(\cdot)$ is max-pooling, and $\mathcal{N}(\cdot)$ denotes the $K^l / 2$ nearest neighbors. 

\noindent\textbf{Textual prompt encoder.} 
Textual prompts provide task-specific instructions to 3DLAs. For 3D dense captioning, we instruct the model to perform one of two tasks: ``describe" or ``describe and localize" the object, while for 3D question answering, we use textual instructions that ask the model to either ``answer" or ``answer and localize the related objects."
Specifically, we encode the input text prompt $\mathcal{I}^t$ using a transformer architecture inspired by BLIP-2~\cite{li2023blip,chen2024ll3da}. This transformer is initialized with a pre-trained BERT model to handle word and positional embeddings, producing text representations $\mathcal{F}^e \in \mathbb{R}^{T \times d_e}$.

\subsection{Multimodal adapter}\label{sec:mma}
Since the 3D and language representations reside in distinct latent spaces, the multimodal adapter (MMA) aims to bridge the gap between outputs of frozen unimodals. MMA aggregates such multimodal information with a fixed set of 32 learnable query tokens.
Specifically, we implement MMA with a Q-Former architecture~\cite{li2023blip} with transformer layers, featuring 12 attention heads per layer. 
In each layer, these queries interact with the encoded visual prompts, $\left[f_{\text{cli}}; f_{\text{box}}\right]$, and the textual instructions, $I_t$, through a shared self-attention mechanism.
Next, the learnable query tokens and visual prompts interact with our detail-enriched scene representation, $\hat{\mathcal{F}}^g$, via cross-attention. The output of the MMA is a set of 32 queries, denoted as $Q \in \mathbb{R}^{32 \times 768}$, which are then projected into the latent space of LLM through a simple linear projector.

\subsection{Next token prediction loss}
We employ standard language modeling conditioned on the text prompt $\mathcal{I}^t$, visual prompt $\mathcal{I}^v$, and point cloud $\mathcal{P}$, to train on a large text corpus $\mathcal{O}_1,\mathcal{O}_2,\cdots,\mathcal{O}_T$ by performing a next-token prediction task.
The goal is to maximize the probability of $\mathcal{O}_{i+1}$ (the next token) conditioned on the sequence of prior tokens $\mathcal{O}_{i:1}=\mathcal{O}_i,\cdots, \mathcal{O}_1$, $\mathcal{I}^t$, $\mathcal{I}^v$ and $\mathcal{P}$.
The learning objective, $\mathcal{L}_{pred}$, minimizes the cross-entropy loss as follows:
\begin{equation}
    \mathcal{L}_{\text{pred}} = -\sum_i \log P_{\theta}\left(\mathcal{O}_{i+1} \mid \mathcal{O}_{1:i};\mathcal{I}^t;\mathcal{I}^v; \mathcal{P}\right),
\end{equation}
where ${\theta}$ represents the learnable parameters of \ourmethod. 

\section{Hilbert-based Serialization and Partition}
\label{supp:hilbert}
\begin{algorithm}[t!]
\small
\caption{Hilbert-based serialization and partition}
\label{alg:hilbert_curve}
\begin{algorithmic}[1]
\REQUIRE Point cloud $\mathcal{P} {=} \{\bm{p}_i {\in} \mathbb{R}^3 {\mid} i {=} 1, 2, \dots, N\}$, resolution $d$.
\ENSURE Hilbert indices $\mathcal{H} = \{h_i \mid i = 1, 2, \dots, N\}$.

\STATE \textbf{Normalize the points:}
$
\bm{p}_i = \frac{\bm{p}_i - \bm{p}_{\text{min}}}{\bm{p}_{\text{max}} - \bm{p}_{\text{min}}}, \quad \forall \bm{p}_i \in \mathcal{P}
$

\STATE \textbf{Discretize the unit cube:}
$
\bm{p}_i^{\text{grid}} = \lfloor \bm{p}_i \cdot 2^d \rfloor
$
where $2^d$ defines the resolution of the grid.

\STATE \textbf{Convert grid indices to binary:} Represent each grid index $(x, y, z)$ with $d$ bits.\\
$t = (b_{t,d-1}, b_{t,d-2}, \dots, b_{t,0}), \quad t\in\{x,y,z\}$
\STATE \textbf{Transform to Gray code:} Convert binary indices to Gray code to ensure spatial locality:
$
g_{i} = b_i \oplus b_{i+1}, \text{ for } i = 0, \dots, d-2
$

\STATE \textbf{Interleave bits (see \cref{alg:interleave_bits}):} Interleave the Gray code bits of $x$, $y$, and $z$ to form a single integer $h_i$: \\
$h_i = \text{Interleave}(g_x, g_y, g_z)$

\STATE \textbf{Apply recursive rotations:} Use the Hilbert curve recursive structure to reorder the interleaved bits, ensuring continuity of the curve.

\STATE \textbf{Output the Hilbert index:} Combine the reordered bits to compute the Hilbert index $h_i$ for each point.

\STATE \textbf{Sort by Hilbert index:} Sort the points $\mathcal{P}$ based on their Hilbert indices $\mathcal{H}$:\\
$
\mathcal{P}_{\text{sorted}} = \text{Sort}(\mathcal{P}, \mathcal{H})
$

\STATE \textbf{Partition the point cloud:} Divide the sorted points into $L$ spatially coherent partitions:
$\mathcal{P}_1, \mathcal{P}_2, \dots, \mathcal{P}_L$,
where each partition contains approximately $N/L$ points.
\end{algorithmic}
\end{algorithm}
\begin{algorithm}[ht!]
\small
\caption{Interleaving bits, \ie, $\text{Interleave}(\cdot)$}
\label{alg:interleave_bits}
\begin{algorithmic}[1]
\REQUIRE Coordinates $x$, $y$, $z \in \mathbb{N}$, each represented with $d$ bits.
\ENSURE Hilbert index $h$.
\STATE Initialize $h \gets 0$.
\FOR{$i = 0$ to $d-1$}
    \STATE Extract the $i$-th bit from $x$, $y$, and $z$ by
    $b_{x,i} = (x \gg i) \& 1, b_{y,i} = (y \gg i) \& 1,  b_{z,i} = (z \gg i) \& 1$.
    \STATE Interleave the bits into $h$: \\
    $h \gets h {\mid} (b_{x,i} {\ll} (3i)) {\mid }(b_{y,i} {\ll} (3i + 1)) \mid (b_{z,i} {\ll} (3i + 2))$.
\ENDFOR
\RETURN $h$
\end{algorithmic}
\end{algorithm}
We choose Hilbert-based serialization for partitioning unordered point clouds for its efficiency and effectiveness in handling large point clouds. 
While grid partitioning is faster ($O(N)$), the Hilbert curve offers superior spatial coherence and is computationally simpler than KD-trees or octrees, making it particularly beneficial for tasks such as clustering and spatial indexing (more details please refer~\cite{moon2001analysis}). 
Additionally, computing point indices with Hilbert-based serialization is inherently parallelizable, as each point can be processed independently. Sorting and partitioning steps can also benefit from such parallel algorithms.

In the following subsections, we describe the Hilbert-based serialization process in detail, and provide analysis on its computational complexity.

\begin{figure*}[t!]
    \centering
    \setlength{\tabcolsep}{0.pt} 
    \renewcommand{\arraystretch}{1.0} 
    \begin{tabular}{cccccc}
        \begin{overpic}[width=0.15\linewidth]{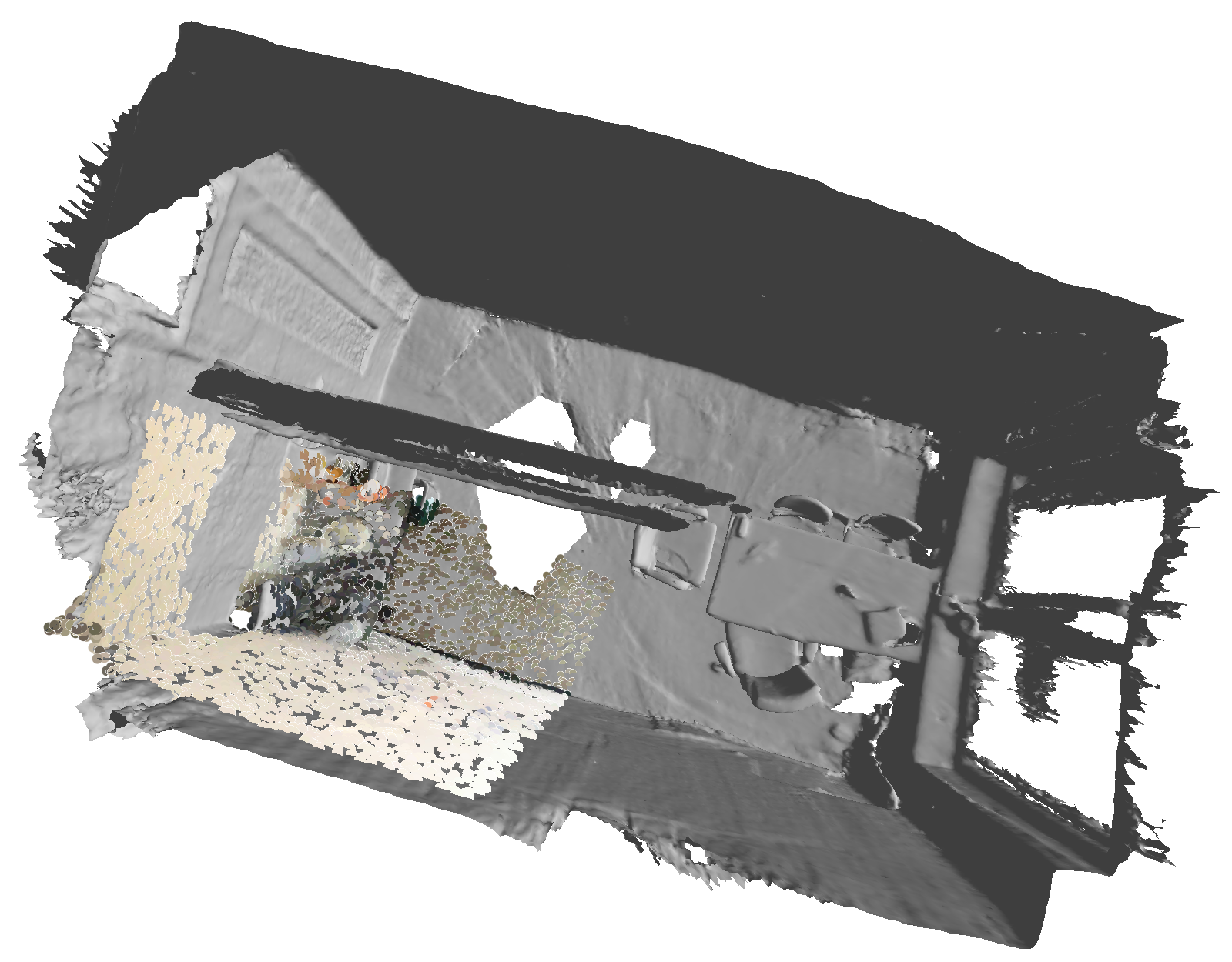}
        \end{overpic} &
        \begin{overpic}[width=0.15\linewidth]{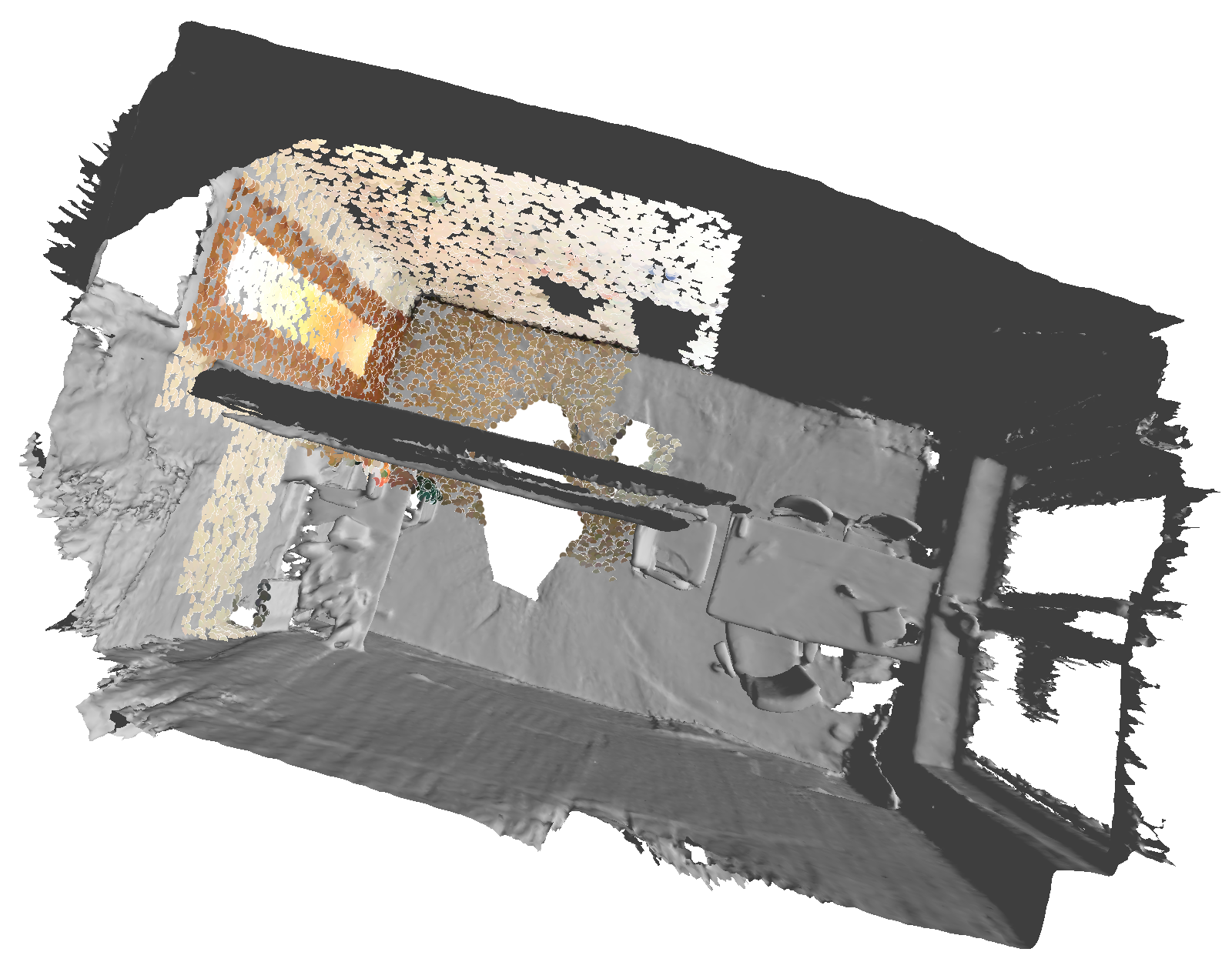}
        \end{overpic} &
        \begin{overpic}[width=0.15\linewidth]{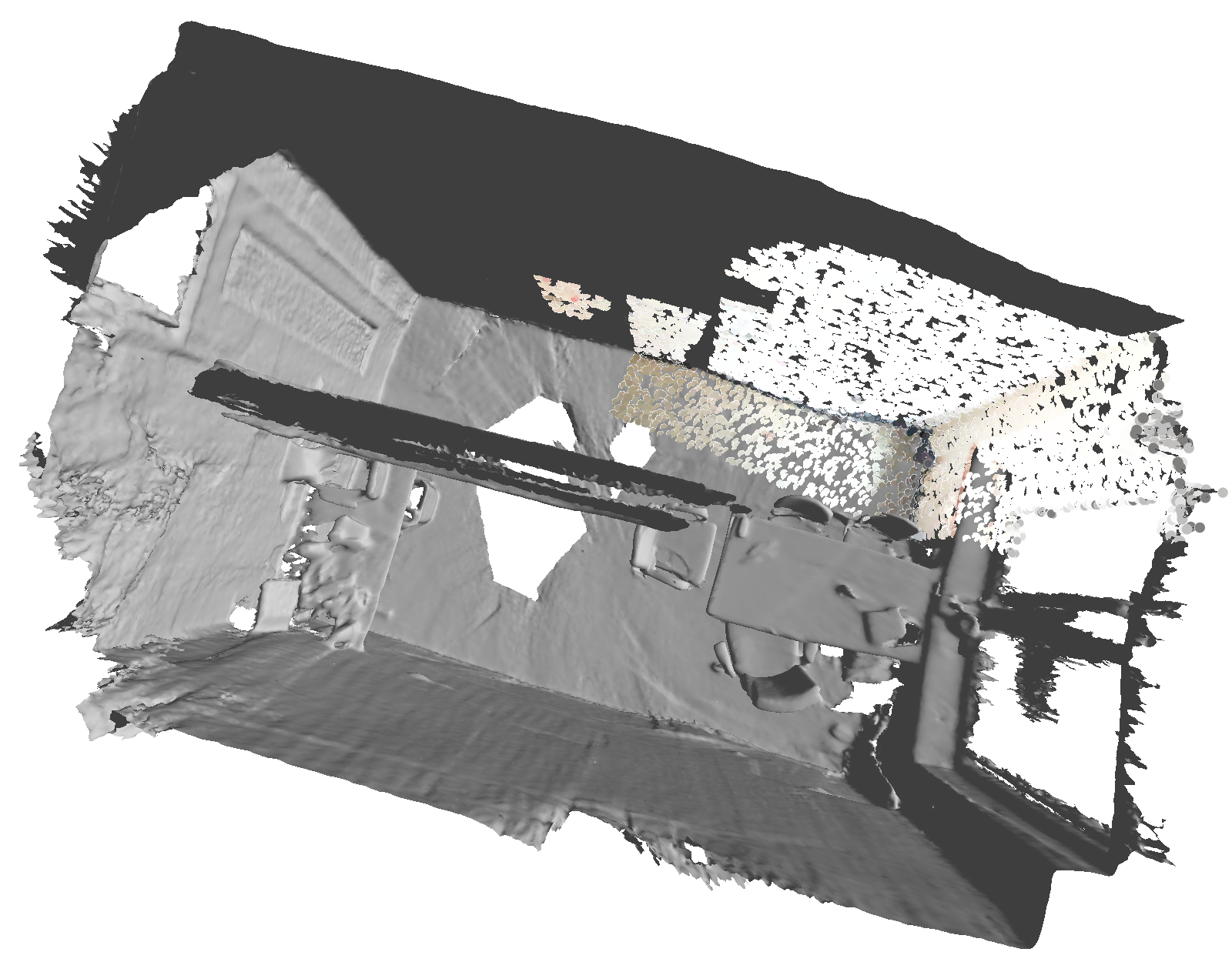}
        \end{overpic} &
        \begin{overpic}[width=0.15\linewidth]{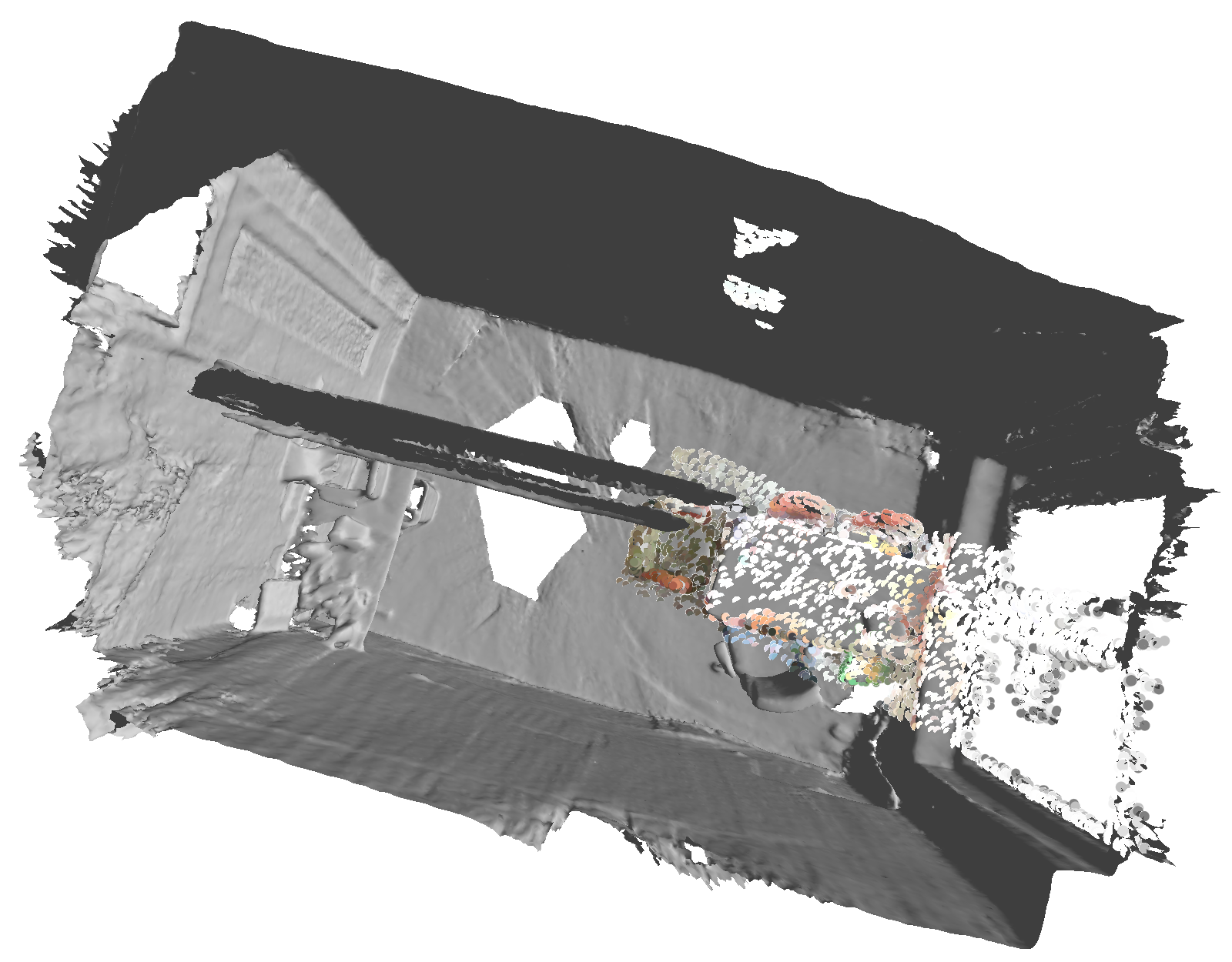}
        \end{overpic} &
        \begin{overpic}[width=0.15\linewidth]{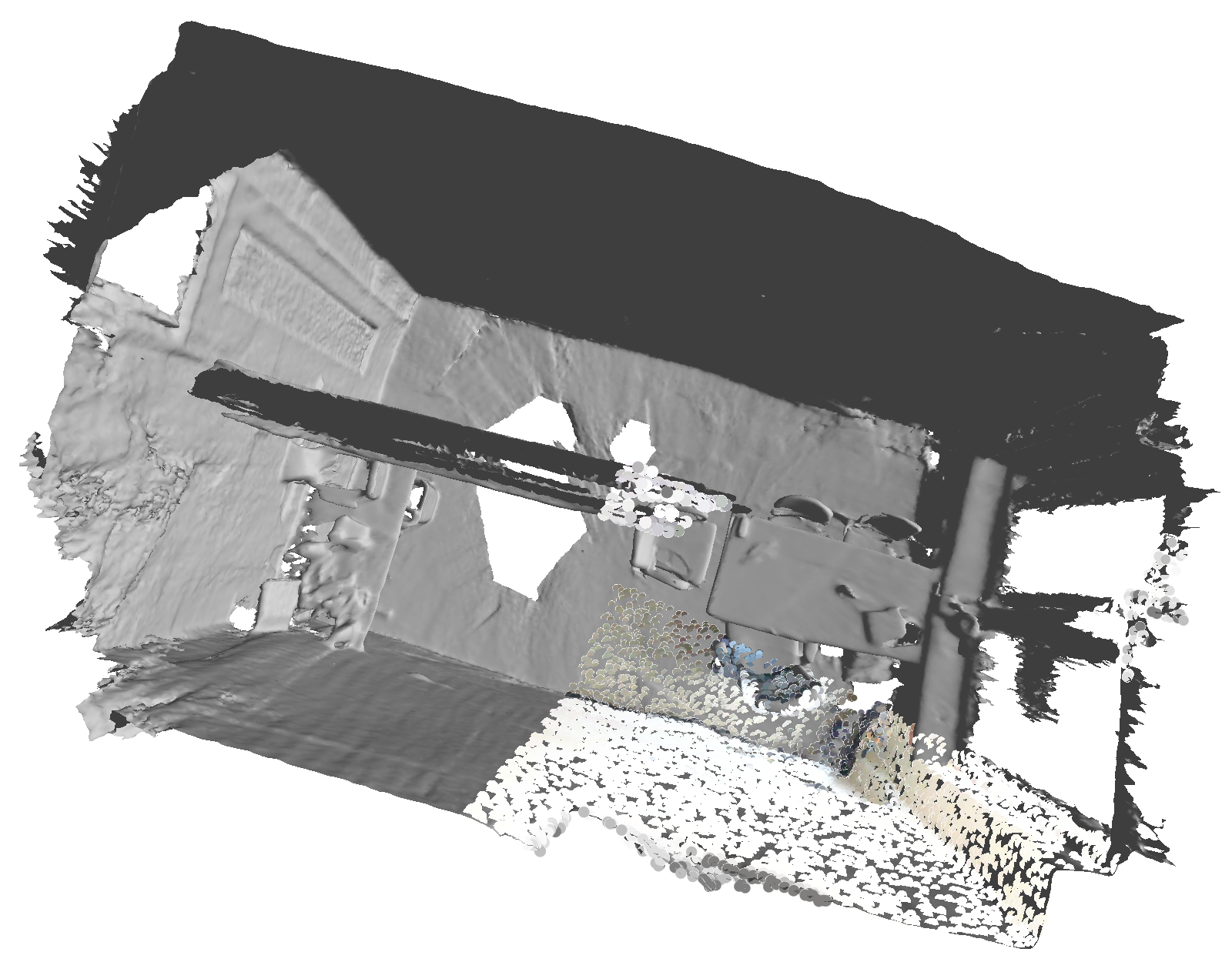}
        \end{overpic} &
        \begin{overpic}[width=0.15\linewidth]{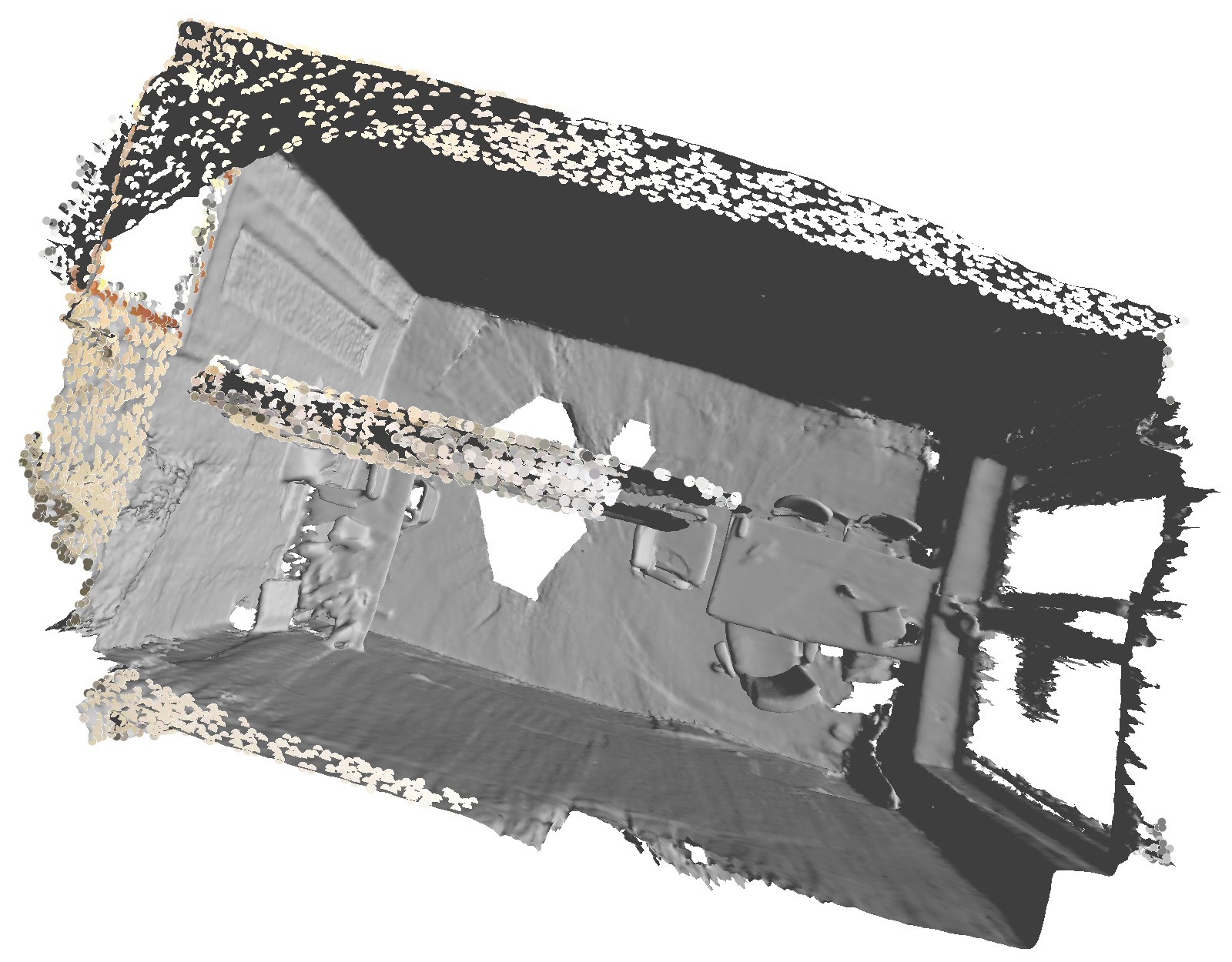}
        \end{overpic} \\
        \begin{overpic}[width=0.15\linewidth]{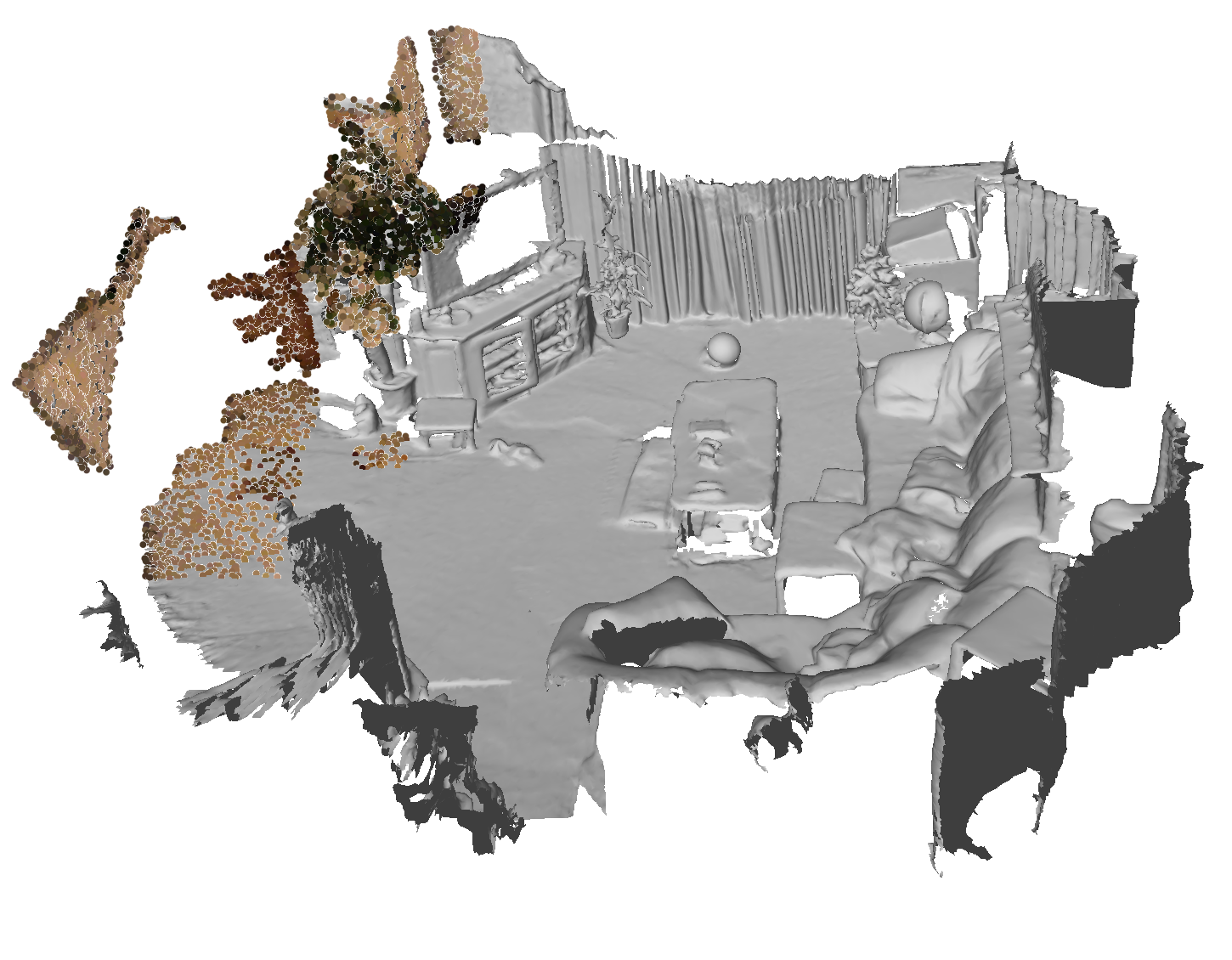}
        \end{overpic} &
        \begin{overpic}[width=0.15\linewidth]{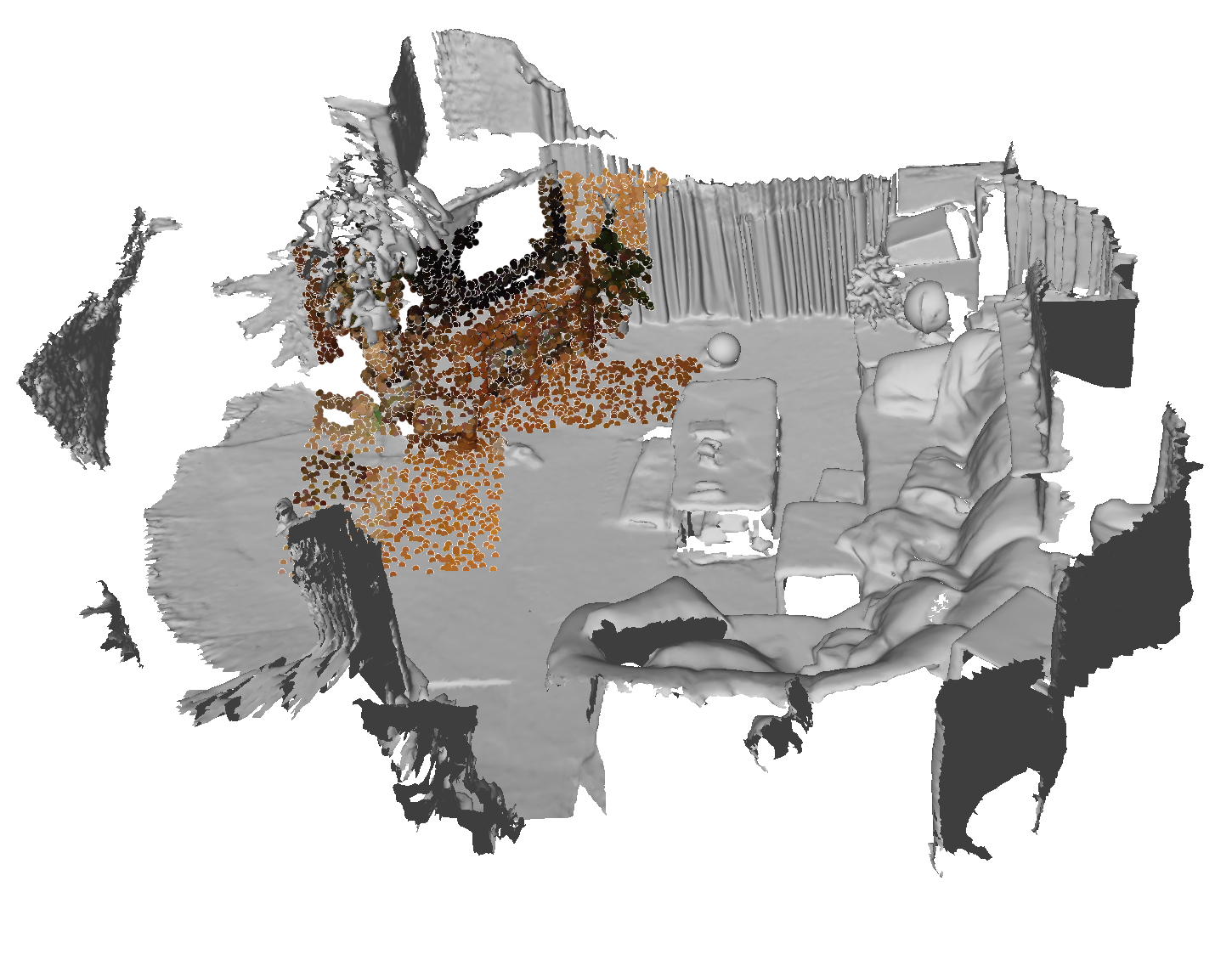}
        \end{overpic} &
        \begin{overpic}[width=0.15\linewidth]{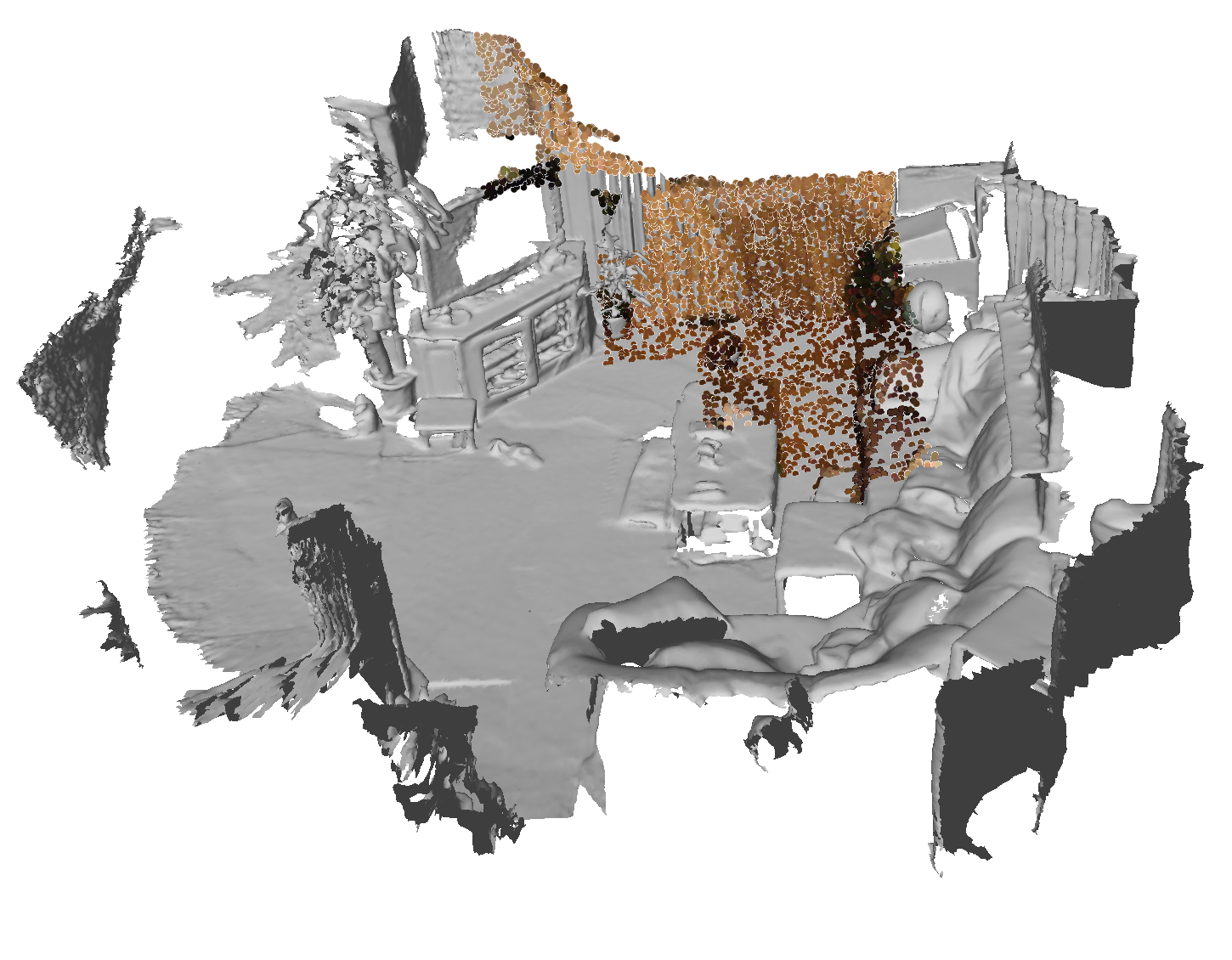}
        \end{overpic} &
        \begin{overpic}[width=0.15\linewidth]{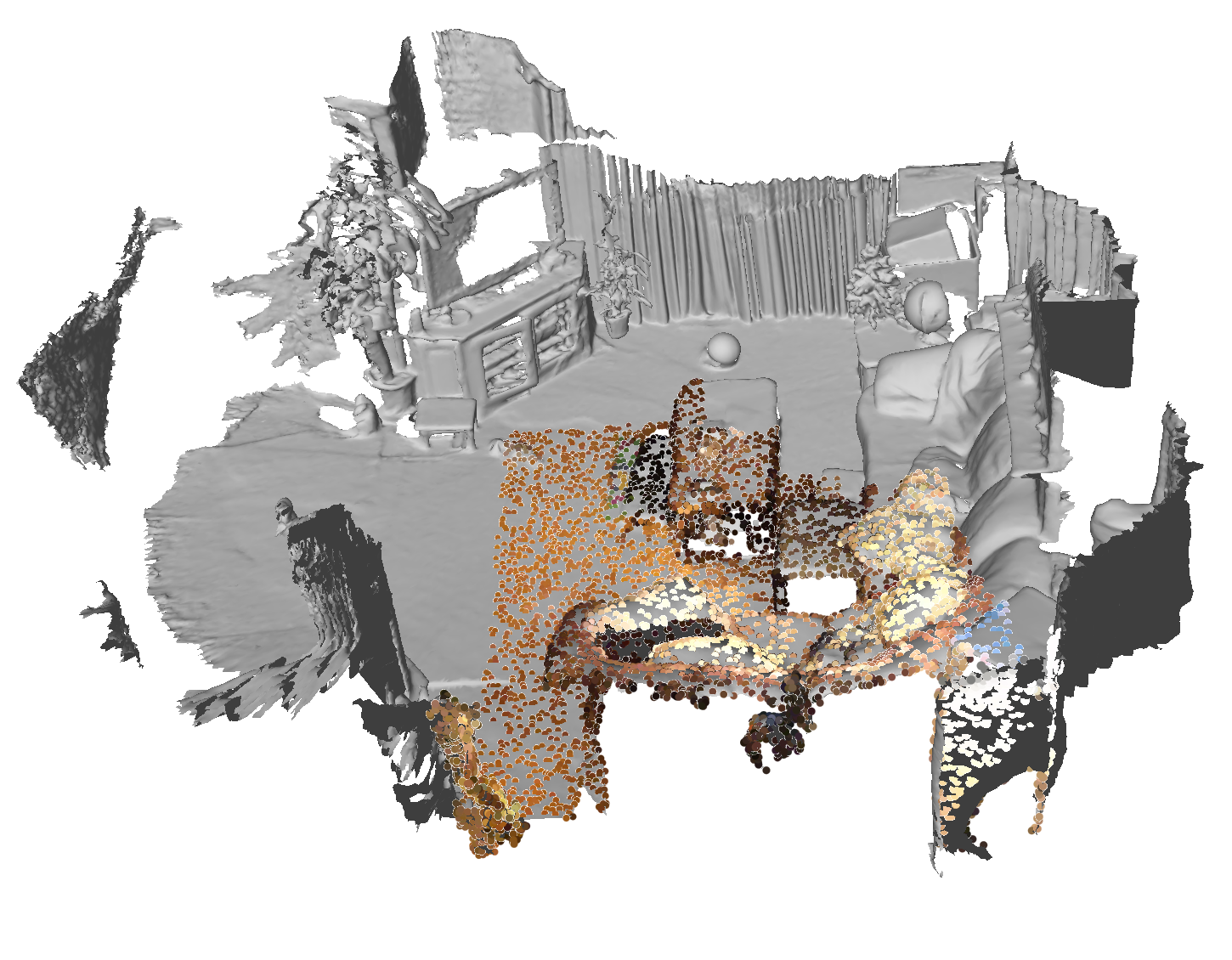}
        \end{overpic} &
        \begin{overpic}[width=0.15\linewidth]{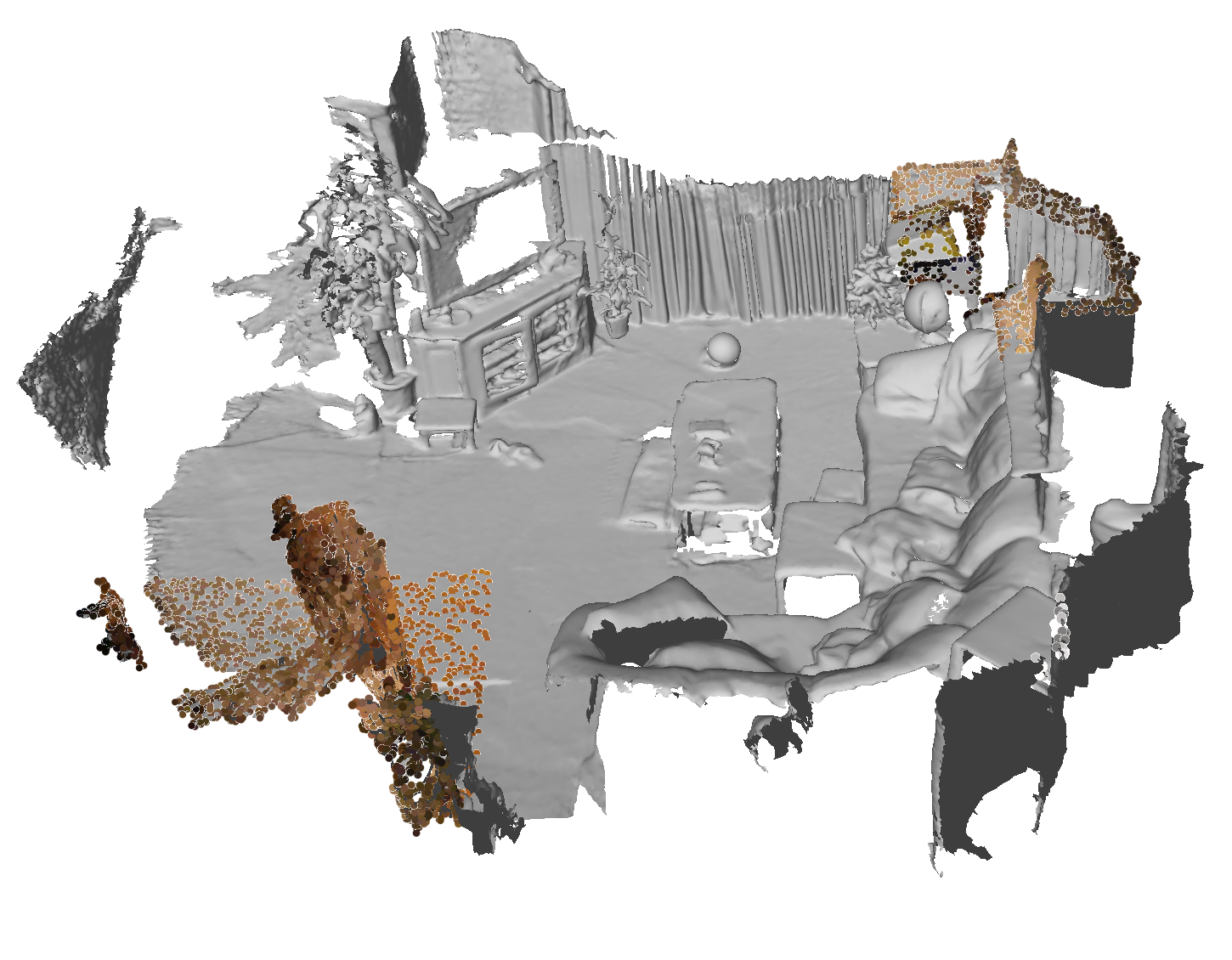}
        \end{overpic} &
        \begin{overpic}[width=0.15\linewidth]{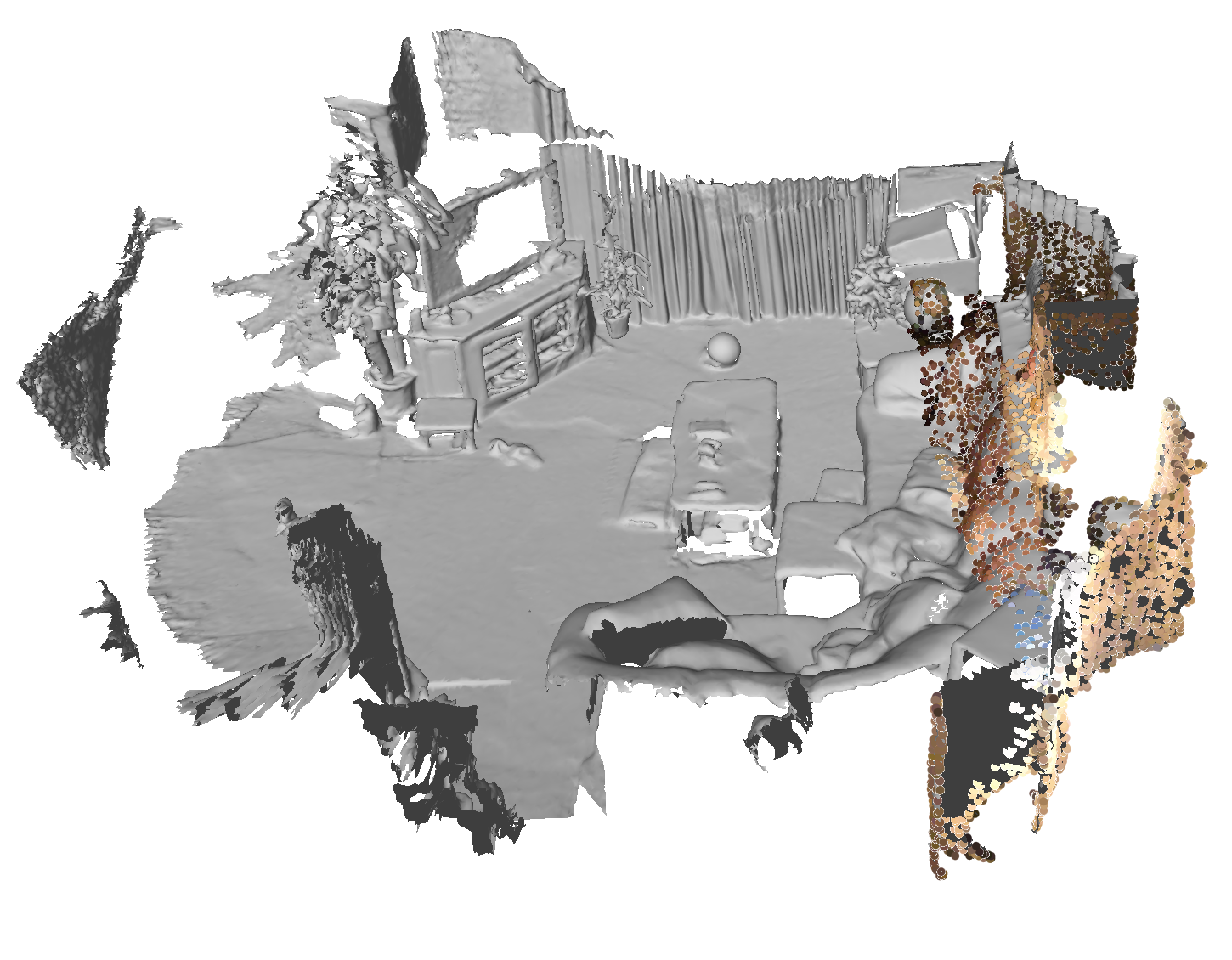}
        \end{overpic} 
    \end{tabular}
    \vspace{-3mm}
    \caption{Visualization of two qualitative examples demonstrating scene partitioning using Hilbert-based serialization. The images illustrate the stepwise refinement of point cloud partitions, with each row corresponding to a different scene example. From left to right, the partitions (highlighted with brownish color) evolve as the serialization method groups spatially adjacent points.}
    \label{fig:sup_partition}
\end{figure*}
\subsection{Algorithm details}
We outline the steps for partitioning a point cloud using the Hilbert curve, as detailed in~\cref{alg:hilbert_curve}. This approach leverages the locality-preserving properties of the Hilbert curve to organize and group points into spatially consistent parts.
The key steps include: normalizing the point cloud to fit within a unit cube, discretizing the unit cube into a grid, calculating Hilbert indices (by converting grid indices to binary, applying Gray code transformation, and interleaving bits), sorting the points based on their Hilbert indices, and partitioning the point cloud accordingly.

\subsection{Computation efficiency}
We analyze the computational efficiency of Hilbert curve partitioning in terms of both time and space complexity:

\noindent\textbf{Time complexity} is $O(N \cdot d + N \log N)$ (dominated by sorting for large $N$) and is influenced by the following key factors: 1) \textit{Mapping Points to Hilbert Curve:} normalizing, binary conversion, Gray code conversion, and bit interleaving take $O(N \cdot d)$, where $N$ is the number of points and $d$ is the resolution in bits.
2) \textit{Sorting:} sorting the Hilbert indices requires $O(N \log N)$ using efficient sorting algorithms.
3) \textit{Partitioning:} dividing the sorted points into $k$ parts is $O(N)$.

\noindent\textbf{Space complexity} is $O(N \cdot d)$, determined by the following factors: 1) storing points and Hilbert indices requires $O(N)$, and 2) binary representations and intermediate data take $O(N \cdot d)$.

\cref{tab:sup_efficiency_comparison} compares the total hard drive usage and processing time of different methods. For 3D-LLM~\cite{hong20233dllm}, multi-view representations are extracted from ScanNet videos every 20 frames, following the protocol outlined in LL3DA~\cite{chen2024ll3da}. The results demonstrate the efficiency of \ourmethod, achieving a good balance between storage requirements and processing time, significantly outperforming 3D-LLM and comparable to LL3DA in computational cost.

\begin{table}[t]
\small
    \centering
    \caption{Computational analysis among 3D-LLM~\cite{hong20233dllm}, LL3DA~\cite{chen2024ll3da}, and \ourmethod.}
    \label{tab:sup_efficiency_comparison}
    \vspace{-3mm}
    \begin{tabular}{l|ccc} 
        \toprule
        \textbf{Cost} & {3D-LLM}~\cite{hong20233dllm} & {LL3DA~\cite{chen2024ll3da}} & \ourmethod \\ 
        \midrule
        Hard Drive (GB)$\downarrow$ & 74563.49 & 5.92 & 41.58 \\ 
        Time per Scene (s)$\downarrow$ & 48203.14 & 6.24 & 13.79 \\ 
        \bottomrule
    \end{tabular}
\end{table}

\noindent\textbf{Hilbert-based nearest-neighbor search.}
Our \knn search search is constrained by geometric segments. We first serialize the union of the global and local point clouds 
$\mathcal{P}^g \cup \mathcal{P}^l$, using a Hilbert curve ordering. This ordering maps each point’s multi-dimensional coordinates to a one-dimensional Hilbert bit $d$, thereby preserving spatial locality. Next, we apply geometric partitioning~\cite{landrieu2018large} on the original point clouds to generate geometric labels $ \mathcal{Y}^g $ and $ \mathcal{Y}^l $.
For each label, we compute its center (\ie, the mean coordinate of all points in that label) and incorporate this center information into the serialized index as high‐order bits via bit shifting. In practice, this is achieved by computing a combined metric for each point:
\begin{equation*}
     \text{combined} = \texttt{label\_offset} \times \text{label} + d,
\end{equation*}
where $\texttt{label\_offset}$ is chosen to be larger than the maximum possible Hilbert distance. 
Once the union is sorted per batch using this combined index, we perform an approximate \knn search.

\subsection{Qualitative examples with partitioning.}
In \cref{fig:sup_partition}, we provide two qualitative examples of point cloud partitioning, achieved through Hilbert-based serialization. This approach groups spatially adjacent points into partitions that preserve locality, effectively encoding spatial relationships within the 3D scenes. The top row depicts the progressive partitioning of the first scene, while the bottom row shows the same process for a second scene. From left to right, the images demonstrate how the Hilbert-based method refines partitions, capturing the hierarchical structure of the point cloud. These examples highlight the ability of Hilbert-based serialization to produce coherent partitions that align with the underlying spatial organization of the scenes.
\section{Dataset Details} \label{supp:dataset}
During the training phase, we leverage the training set of the ScanNet portion from the 3DLLM dataset~\cite{hong20233dllm}. Additionally, we incorporate data from complementary datasets, including ScanQA~\cite{azuma2022scanqa}, ScanRefer~\cite{chen2020scanrefer}, and Nr3D~\cite{achlioptas2020referit3d}. The dataset details are provided below.

\noindent\textbf{3D-LLM dataset~\cite{hong20233dllm}} comprises: i) 1,033 textual descriptions across 517 scenes, ii) 1,864 lines of embodied task planning spanning 510 scenes, and iii) 2,955 lines of multi-turn embodied dialogues across 517 scenes.

\noindent\textbf{ScanQA dataset~\cite{azuma2022scanqa}} is a 3D question-answering benchmark built on top of the ScanRefer~\cite{chen2020scanrefer} dataset, designed to evaluate the ability of models to understand 3D scenes through natural language queries. The dataset contains 6,857 unique questions paired with 30,769 answers spanning 806 reconstructed indoor environments from ScanNet~\cite{dai2017scannet}. Each question focuses on objects within the scene, addressing a variety of topics such as object attributes, spatial relationships, and scene semantics. On average, each scene contains 8.5 questions, encouraging models to reason about object-level details and contextual relationships within complex 3D environments.

\noindent\textbf{ScanRefer dataset~\cite{chen2020scanrefer}} is a 3D language grounding benchmark built on the ScanNet~\cite{dai2017scannet} dataset, consisting of 1,613 RGB-D scans across 806 unique indoor environments. The dataset provides natural language descriptions for objects in reconstructed 3D scenes, with a total of 51,583 descriptions covering 800 ScanNet scenes. Each object is annotated with an average of 4.67 descriptions, ensuring comprehensive linguistic diversity. On average, each scene contains 13.81 objects and 64.48 descriptions, spanning over 250 types of common indoor objects. Among these, 41,034 descriptions explicitly mention object attributes such as color, shape, size, and spatial relationships, making the dataset a rich resource for evaluating fine-grained language grounding in complex 3D environments.

\paragraph{Nr3D dataset~\cite{achlioptas2020referit3d}} is a benchmark for 3D object localization tasks in natural language, built on the ScanNet~\cite{dai2017scannet} dataset. It contains 41,503 unique natural language descriptions referring to 5,578 objects across 707 ScanNet scenes. Each description is designed to unambiguously identify a target object in the context of its surrounding scene, incorporating spatial relationships and object attributes such as color, shape, and size. On average, each object is associated with 7.4 descriptions, providing comprehensive linguistic diversity. The dataset focuses on common indoor objects, making it suitable for evaluating fine-grained understanding of object attributes and spatial reasoning in 3D scenes.

\section{Additional analysis}\label{supp:analysis}
\subsection{Details with increasing tokens (LL3DA$^\dag$)}
In the main paper (Tab.~3), we introduce a variant of LL3DA$^\dag$, which combines both global and local information by leveraging query tokens generated from local and global regions. LL3DA$^\dag$ serves as a baseline of enriching global context with local details by extending the number of query tokens. 
LL3DA$^\dag$ first extracts 3D representations from different partitions of the scene. It then applies Farthest Point Sampling (FPS) to select 1,024 points, along with their corresponding point-level representations, from the union of these partitions. These sampled representations are processed through the multimodal adapter to produce 32 local tokens. Then, we can generate an additional 32 tokens (global tokens) from the point cloud of the entire scene. We finally obtain a total of 64 tokens by concatenating such local and global tokens and processed through a self-attention layer, \ie, $\mathcal{Q}^{64\times 768} = f_{att}\left(\mbox{cat}[\mbox{MMA}(\mathcal{F}^g,\mathcal{I}^t), \mbox{MMA}(\mathcal{F}^l,\mathcal{I}^t)]\right)$ (where the number of tokens increases from 32 to 64). The self-attention mechanism enables interaction and information exchange between the 64 tokens, enhancing the representation of both global and local representations.
We evaluate the performance of LL3DA$^\dag$ on the ScanQA validation dataset, comparing it with the original LL3DA and our \ourmethod{}. As shown in~\cref{tab:sub_ab_2xtoken}, increasing the number of tokens enables LL3DA$^\dag$ to achieve a modest performance gain over the reproduced LL3DA (repr.), as expected.
However, its effectiveness remains limited compared to \ourmethod{}. This is because \ourmethod{} is specifically designed to capture both global context and local details during the scene encoding phase. 
Furthermore, it is trained with carefully crafted loss functions that ensure stable and efficient learning.
Our findings align with Idefics2~\cite{laurenccon2024matters}. 
Idefics2 shows that reducing the number of visual tokens through attention-based pooling significantly enhances computational efficiency during training and inference while improving performance on downstream tasks.


\begin{table}[t] 
\centering 
\small
\caption{Ablation study on the impact of the increasing the number tokens on the ScanQA~\cite{azuma2022scanqa} validation dataset.} \label{tab:sub_ab_2xtoken} 
\vspace{-3mm} 
{%
\begin{tabular}{l|cccc} 
\toprule Method & C$\uparrow$ & B4$\uparrow$ & M$\uparrow$ & R$\uparrow$ \\
\midrule 
LL3DA (repr.) & 74.37 & 13.50 & 15.09 & 36.31 \\
LL3DA$^\dag$ & 74.54 & 12.89 & 15.11 & 36.96 \\
\rowcolor{lightgreen} \ourmethod & \bf78.13 & \bf14.49 & \bf17.44 & \bf39.60 \\
\bottomrule 
\end{tabular} } 
\vspace{-2mm} 
\end{table} 

\subsection{Scene-aware visual prompt encoder}
In the visual prompt encoder, we integrate the detail-enriched scene representations to enhance the prompt representations for clicks and object bounding boxes as explained in \cref{supp:components}. 
To evaluate the efficiency of the scene-aware visual prompt encoder, we conducted experiments on the ScanQA~\cite{azuma2022scanqa} validation dataset. As shown in \cref{tab:sub_ab_prompt}, by integrating scene representations with the visual prompts, while not being a major contributor to the performance, it brings consistent marginal improvements across all evaluation metrics.

\begin{table}[t] 
\centering 
\small
\caption{Ablation study on the impact of the scene-aware prompt encoder on the ScanQA~\cite{azuma2022scanqa} validation dataset.} \label{tab:sub_ab_prompt} 
\vspace{-3mm} 
{%
\begin{tabular}{l|cccc} 
\toprule 
Method & C$\uparrow$ & B4$\uparrow$ & M$\uparrow$ & R$\uparrow$ \\
\midrule 
without scene-aware & 78.01 & 14.46 & 17.32 & 39.46 \\
\rowcolor{lightgreen} with scene-aware & \bf78.13 & \bf14.49 & \bf17.44 & \bf39.60 \\
\bottomrule 
\end{tabular} } 
\vspace{-2mm} 
\end{table} 

\input{main/tables/ta_sup_ab_token}
\subsection{Increase number of tokens}
To evaluate the impact of the number of learnable query tokens, we experimented with various numbers of learnable query tokens, from 32 to 128, in the Q-Former (different from LL3DA$^\dag$).
We trained each model \textit{from scratch on the ScanQA training set} and evaluated its performance on the validation set for both LL3DA and \ourmethod{}.
We observed that increasing the number of tokens to 96 and 128 led to loss divergence (to infinity) in both LL3DA and \ourmethod{}.
To address this instability, we incorporated mirror descent-based regularization~\cite{wang2019divergence}.
\cref{sup:tab_token} presents the results.
Both LL3DA and \ourmethod achieve better performance with increasing query tokens, where the performance gain tends to saturate from 96 to 128 tokens.
Moreover, our proposed method \ourmethod consistently outperforms LL3DA across all token configurations and evaluation metrics.


\subsection{Performance for two stage training} 
\input{main/tables/tab_gene}
\input{main/tables/tab_sup_3dllm}
As in prior works~\cite{chen2024ll3da,hong20233dllm}, \ourmethod uses a two-stage training strategy, where the model is pre-trained on an ensemble dataset comprising diverse 3D tasks. This ensemble dataset allows the model to develop a broad understanding of various 3D scenarios, building the base as a 3D \textit{generalist} model, such as scene description, dense captioning, and question answering. 
Then, instruction-following fine-tuning is further applied to the generalist model to further enhance its performance on specialized downstream tasks, such as 3D dense captioning and 3D question answering, using task-specific datasets.

We evaluate the performance of \ourmethod~under this two-stage training paradigm, as summarized in \cref{tab:two_stage}, together with several baseline methods.
The first three rows in \cref{tab:two_stage} display the performance of models trained from scratch as task-specific experts. The next three rows show the results of models fine-tuned on individual tasks, initialized from the generalist model's weights. The final row reports the performance of the generalist model without fine-tuning, where a single set of weights is used to handle all tasks.

Our generalist model (the last row without fine-tuning) demonstrates strong task differentiation capabilities, excelling in tasks such as 3D dense captioning and 3D question answering when provided with appropriate textual instructions and visual prompts. For instance, the fine-tuned model achieves notable improvements on ScanRefer (69.41 C@0.5) and ScanQA (78.13 CiDEr), showcasing its ability to leverage the generalist pre-training to boost downstream task performances.
However, the generalist model exhibits relatively lower performance on Nr3D compared to ScanRefer, likely because these datasets address the same dense captioning task, and explicit differentiation between them was not included during training. Despite this, the model achieves higher scores on ScanRefer (69.41 C@0.5), potentially indicating a preference for dataset-specific characteristics or structural differences between the datasets.
Importantly, the generalist weights serve as a strong initialization for fine-tuning specific tasks. For example, the fine-tuned model on ScanRefer reaches 69.41 C@0.5, significantly outperforming the model trained from scratch (63.02 C@0.5). This highlights the advantages of pretraining as a generalist in boosting task-specific performance.
Overall, the results demonstrate that such two-stage training approach enables effective multi-task learning while maintaining robust performance across individual tasks, even when faced with diverse datasets and objectives. 

\begin{figure*}[t]
    \centering
    \begin{overpic}[width=1\linewidth]{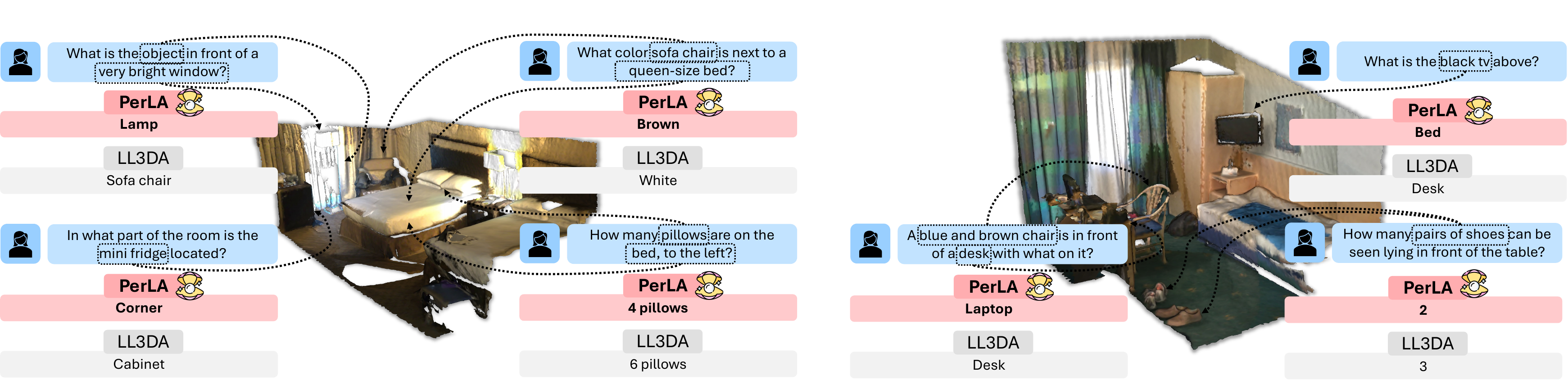}
        \put(40,23){\footnotesize{(a) 3D question answering on ScanQA~\cite{azuma2022scanqa}}}
    \end{overpic}
    \vspace{2mm}
    \begin{overpic}[width=1\linewidth]{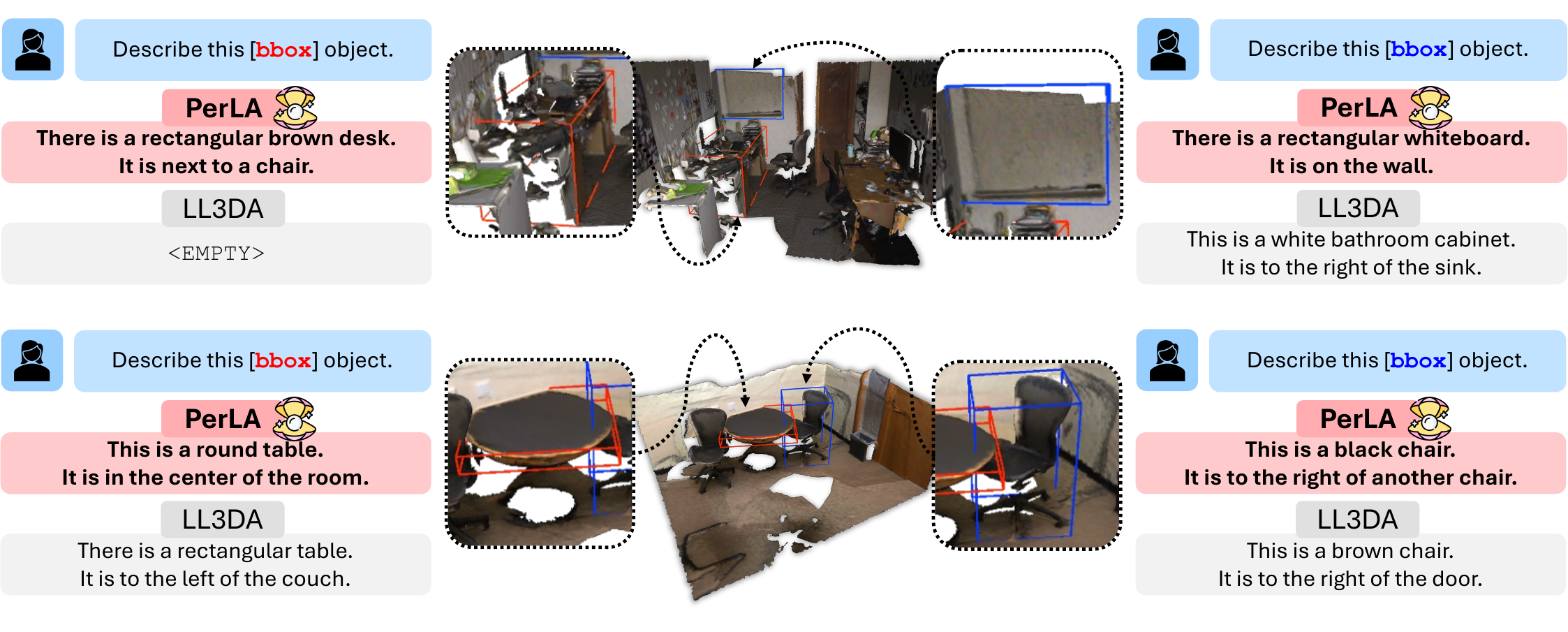}
        \put(40,38){\footnotesize{(b) 3D dense captioning on ScanRefer~\cite{chen2020scanrefer}}}
        \put(40,20){\footnotesize{(b) 3D dense captioning on ScanRefer~\cite{chen2020scanrefer}}}
    \end{overpic}
    \vspace{2mm}
    \begin{overpic}[width=1\linewidth]{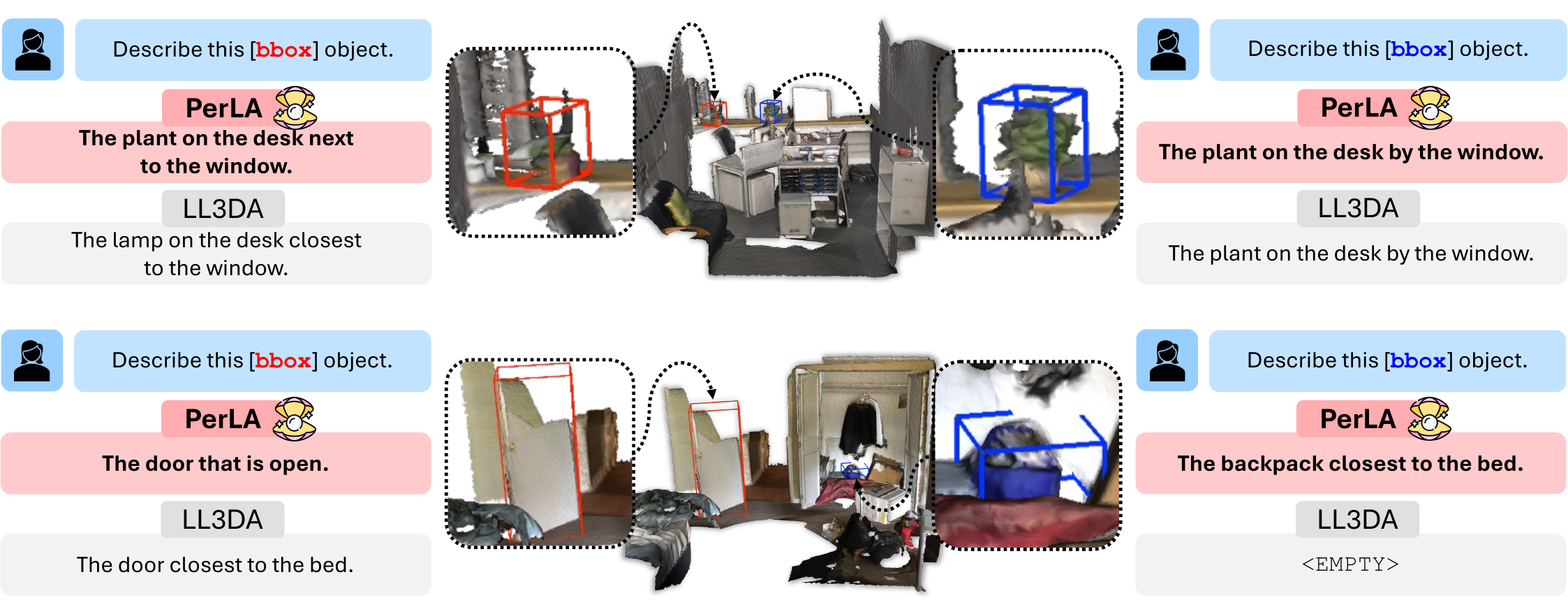}
        \put(40,38){\footnotesize{(d) 3D dense captioning on Nr3D~\cite{achlioptas2020referit3d}}}
        \put(40,18){\footnotesize{(e) 3D dense captioning on Nr3D~\cite{achlioptas2020referit3d}}}
    \end{overpic}
    \caption{Qualitative results for 3D scene understanding tasks, including (a) question answering on ScanQA~\cite{azuma2022scanqa}, (b,c) dense captioning on ScanRefer~\cite{chen2020scanrefer}, and (d,e) dense captioning on Nr3D~\cite{achlioptas2020referit3d}. 
    On ScanQA, \ourmethod successfully identifies and reasons about objects and their relationships in the scene. For example, when asked, ``What is the color of the sofa chair next to a queen-sized bed?", \ourmethod accurately answers by localizing the relevant chair and determining its color. Similarly, for complex spatial queries like ``Which item is to the left of the bookshelf? \ourmethod shows a clear understanding of spatial relationships, providing correct and concise answers. For ScanRefer, \ourmethod demonstrates robust descriptive capabilities by capturing object attributes (e.g., ``the rectangular brown desk" and ``the round table in the center of the room") and spatial relationships. Compared to LL3DA~\cite{chen2024ll3da}, which often generates incomplete or erroneous descriptions, \ourmethod excels in producing detailed and accurate outputs. Similarly, on Nr3D, \ourmethod showcases fine-grained spatial reasoning, with outputs like ``the door that is open" and ``the backpack closest to the bed," emphasizing its superior understanding of object attributes and spatial relationships.
    }
    \label{fig:sup_dense_caption}
\end{figure*}

\subsection{More results on Scene Description, Embodied Dialogue and Embodied Planning}
To provide a comprehensive evaluation, we present additional results on the tasks of scene description, embodied dialogue, and embodied planning using the ScanNet subset of the 3D-LLM dataset~\cite{hong20233dllm}, which has been used as part of our pre-training dataset. Consistent with the dataset split defined prior work~\cite{chen2024ll3da}, scenes with IDs less than 600 are used for training, while the remaining scenes are reserved for validation.
We evaluate these tasks using the same metrics as in the main paper: BLEU-n~\cite{papineni2002bleu} (1-4), CiDER~\cite{vedantam2015cider}, METEOR~\cite{banerjee2005meteor}, and Rouge-L~\cite{lin2004rouge}. 

As shown in~\cref{tab:scene_comparisons}, for scene description, \ourmethod~outperforms all baselines, including LL3DA, across all metrics. Notably, it achieves substantial improvements in CiDER (+3.07) and METEOR (+1.54) compared to LL3DA.

For embodied dialogue, while LL3DA has already scored quite competitive performance, \ourmethod achieves further improvements across all metrics, with significant margins in BLEU-4 (+1.34), CiDER (+2.66), and METEOR (+6.05). In embodied planning, \ourmethod achieves the highest scores across all metrics, surpassing LL3DA, with notable improvements in CiDER (+16.93) and METEOR (+6.41)

These results affirm that \ourmethod, by enhancing the capability of 3DLAs in perceiving 3D scene details, can consistently improve performance on various downstream tasks, being beneficial to not only perception tasks but also tasks in general embodied context, such as planning. 


\subsection{Qualitative results}
We provide additional qualitative visualization results (\cref{fig:sup_dense_caption}) on two 3D scene understanding tasks: 3D question answering and 3D dense captioning. The visualization includes (a) question answering on ScanQA~\cite{azuma2022scanqa}, (b) dense captioning on ScanRefer~\cite{chen2020scanrefer}, and (c) dense captioning on Nr3D~\cite{achlioptas2020referit3d}. These examples underscore the effectiveness of \ourmethod in producing more accurate responses when addressing questions specific to a given 3D scene.

For the 3D question aswering task on ScanQA, \ourmethod{} is able to correctly interpret natural language questions and respond with accurate answers that are grounded in the 3D scene. For example, when asked, ``What color is the chair near the desk?" or ``How many doors are in the room?", \ourmethod{} can accurately identify relevant objects and their attributes. While in contrast, baseline models like LL3DA~\cite{chen2024ll3da} struggle with questions requiring multi-step reasoning or fine-grained scene comprehension, often producing incomplete or incorrect answers. 
%

For the 3D dense captioning task, \ourmethod{} also demonstrates exceptional descriptive capabilities, accurately capturing object attributes (e.g., ``the rectangular brown desk" and ``the round table in the center of the room") and spatial relationships. In comparison, LL3DA~\cite{chen2024ll3da} might produce incomplete or inaccurate descriptions. \ourmethod’s ability to provide accurate output is particularly evident in challenging cases, such as ``There is a rectangular whiteboard. It is on the wall.'' in ScanRefer, and small or partially reconstructed objects, such as ``The plant on the desk next to the window." in Nr3D. 
Moreover, in the example of ``The door that is open.” and ``The backpack closest to the bed.'' in Nr3D, \ourmethod showcases its advanced fine-grained spatial understanding of the object.
generating outputs such as ``the door that is open” and ``the backpack closest to the bed." These examples highlight its superior understanding of both object attributes and spatial relationships, further distinguishing \ourmethod as a more percepti solution for 3D scene understanding tasks.

%% file: main/tables/ta_sup_ab_token.tex
\begin{table*}[t]
    \tabcolsep 6pt
    \small
    \centering
    \caption{Ablation study on the impact of the numbers of learnable query tokens on the ScanQA~\cite{azuma2022scanqa} validation dataset.}\label{sup:tab_token}
     \vspace{-4mm}
    {%
    \begin{tabular}{l|cccc|cccc|cccc|cccc} 
        \toprule
        Method & \multicolumn{4}{c}{32} & \multicolumn{4}{c}{64} & \multicolumn{4}{c}{96}  & \multicolumn{4}{c}{128}\\
        \midrule
        Metric & 
        C$\uparrow$ & B4$\uparrow$ & M$\uparrow$ & R$\uparrow$ & C$\uparrow$ & B4$\uparrow$ & M$\uparrow$ & R$\uparrow$ & C$\uparrow$ & B4$\uparrow$ & M$\uparrow$ & R$\uparrow$ & C$\uparrow$ & B4$\uparrow$ & M$\uparrow$ & R$\uparrow$\\ 
        \midrule
        LL3DA                           & 73.2 & 12.8 & 14.9 & 36.0 & 73.7 & 13.6 & 15.2 & 35.5 & 74.2 & 13.8 & 15.1 & 36.1 & 74.6 & 13.5 & 15.2 & 36.0 \\ 
        \rowcolor{lightgreen}\ourmethod & 74.4 & 13.7 & 15.8 & 36.5 & 75.1 & 14.2 & 15.9 & 36.6 & 76.4 & 14.3 & 16.3 & 40.0 & 77.0 & 14.3 & 16.4 & 38.2\\
        \bottomrule
    \end{tabular}
    }
    \vspace{-4mm}
\end{table*}

%% file: main/tables/tab_gene.tex
\begin{table*}[t]
\small
\centering
\tabcolsep 6pt
\caption{Performance for Two-Stage Training. The first three rows report the performance of \ourmethod trained from scratch as task-specific experts on their respective training datasets. The subsequent three rows present the results of models fine-tuned on each dataset using weights initialized from the generalist model trained across all task-specific datasets. The final row evaluates the generalist model's performance without fine-tuning. ScanRefer~\cite{chen2020scanrefer} and Nr3D~\cite{achlioptas2020referit3d} are evaluated for dense captioning, while ScanQA~\cite{azuma2022scanqa} is evaluated for question answering. The results demonstrate the effectiveness of our generalist model in multi-task scenarios and its strong performance after fine-tuning.}\label{tab:two_stage}
\vspace{-3mm}
\begin{tabular}{lcccccccccccc}
\toprule
{Method} & \multicolumn{4}{c}{{ScanRefer@0.5}} & \multicolumn{4}{c}{{Nr3D@0.5}} & \multicolumn{4}{c}{{ScanQA}} \\
\cmidrule(lr){2-5} \cmidrule(lr){6-9} \cmidrule(lr){10-13}
 & {C$\uparrow$} & {B4$\uparrow$} & {M$\uparrow$} & {R$\uparrow$} & {C$\uparrow$} & {B4$\uparrow$} & {M$\uparrow$} & {R$\uparrow$} & {C$\uparrow$} & {B4$\uparrow$} & {M$\uparrow$} & {R$\uparrow$} \\
\midrule
ScanRefer (scratch) & 63.02 & 35.02 & 25.61 & 54.09 & - & - & - & - & - & - & - & - \\
Nr3D (scratch) & - & - & - & - & 48.37 & 28.36 & 25.72 & 55.97 & - & - & - & - \\
ScanQA (scratch) & - & - & - & - & - & - & - & - & 74.44 & 13.69 & 15.78 & 36.49 \\
\midrule
ScanRefer (fine-tuned) & \bf69.41 & \bf38.02 & \bf29.07 &  \bf56.80 & - & - & - & - & - & - & - & - \\
Nr3D (fine-tuned) & - & - & - & - & \bf55.06 & \bf31.24 & \bf28.52 & \bf59.13 & - & - & - & - \\
ScanQA (fine-tuned) & - & - & - & - & - & - & - & - & \bf78.13 & \bf14.49 & \bf17.44 & \bf39.60 \\
\midrule
w/o fine-tuning & 66.10 & 37.03 & 27.33 & 54.92 & 51.57 & 28.40 & 26.24 & 56.32 & 76.88 & 14.05 & 16.07 & 37.97 \\
\bottomrule
\end{tabular}
\end{table*}

%% file: main/tables/tab_sup_3dllm.tex
\begin{table*}[t!]
    \centering
    \small
    \caption{Quantitative Comparisons on Scene Description, Embodied Dialogue, and Embodied Planning on the ScanNet part of 3D-LLM~\cite{hong20233dllm} with a beam size = 4 for beam search.}
    \label{tab:scene_comparisons}
    \begin{tabular}{l|l|ccccccc} 
        \toprule
        Task & Method & BLEU-1$\uparrow$ & BLEU-2$\uparrow$ & BLEU-3$\uparrow$ & BLEU-4$\uparrow$ & CiDER$\uparrow$ & METEOR$\uparrow$ & Rouge-L$\uparrow$ \\ 
        \midrule 
        \multirow{5}{*}{\rotatebox{45}{Scene Description}} 
        & OPT-1.3B~\cite{zhang2022opt} & 15.79 & 6.10 & 2.07 & 0.84 & 0.00 & 8.40 & 11.70 \\ 
        & OPT-2.7B~\cite{zhang2022opt} & 19.97 & 7.59 & 3.62 & 1.13 & 0.00 & 6.60 & 12.32 \\ 
        & OPT-6.7B~\cite{zhang2022opt} & 24.40 & 9.93 & 3.64 & 1.13 & 0.06 & 8.99 & 16.96 \\ 
        & LLAMA-7B~\cite{touvron2023llama} & 19.26 & 7.69 & 2.79 & 0.92 & 0.20 & 7.00 & 12.31 \\ 
        & LL3DA~\cite{chen2024ll3da} & 29.94 & 21.56 & 14.93 & 10.02 & 1.32 & 12.31 & 27.08 \\ 
        \rowcolor{lightgreen}& \ourmethod & \bf31.29 & \bf23.67 & \bf16.23 & \bf12.14 & \bf4.39 & \bf13.85 & \bf28.79 \\ 
        \midrule
        \multirow{5}{*}{\rotatebox{45}{Embodied Dialogue}} 
        & OPT-1.3B~\cite{zhang2022opt} & 2.44 & 1.05 & 0.46 & 0.23 & 0.31 & 5.62 & 4.83 \\ 
        & OPT-2.7B~\cite{zhang2022opt} & 3.88 & 1.56 & 0.73 & 0.39 & 0.38 & 7.38 & 6.28 \\ 
        & OPT-6.7B~\cite{zhang2022opt} & 3.59 & 1.65 & 0.81 & 0.43 & 0.25 & 6.88 & 6.16 \\ 
        & LLAMA-7B~\cite{touvron2023llama} & 4.08 & 1.80 & 0.90 & 0.50 & 0.27 & 7.81 & 6.68 \\ 
        & LL3DA~\cite{chen2024ll3da} & 48.14 & 39.83 & 34.83 & 31.32 & 260.07 & 27.21 & 47.69 \\ 
        \rowcolor{lightgreen}& \ourmethod & \bf49.91 & \bf41.10 & \bf36.06 & \bf32.66 & \bf262.73 & \bf33.26 & \bf48.24 \\ 
        \midrule
        \multirow{5}{*}{\rotatebox{45}{Embodied Planning}} 
        & OPT-1.3B~\cite{zhang2022opt} & 1.26 & 0.59 & 0.26 & 0.13 & 0.16 & 0.24 & 3.56 \\ 
        & OPT-2.7B~\cite{zhang2022opt} & 2.02 & 0.99 & 0.49 & 0.26 & 0.10 & 3.59 & 4.35 \\ 
        & OPT-6.7B~\cite{zhang2022opt} & 2.03 & 1.06 & 0.53 & 0.28 & 0.00 & 3.65 & 3.94 \\ 
        & LLAMA-7B~\cite{touvron2023llama} & 2.24 & 1.13 & 0.55 & 0.29 & 0.04 & 3.53 & 4.71 \\ 
        & LL3DA~\cite{chen2024ll3da} & 45.07 & 33.04 & 24.96 & 19.15 & 196.78 & 19.87 & 45.58 \\ 
        \rowcolor{lightgreen}& \ourmethod & \bf48.96 & \bf36.19 & \bf27.82 & \bf22.42 & \bf213.71 & \bf26.28 & \bf47.57 \\ 
        \bottomrule
    \end{tabular}
\end{table*}